%% file: esc23_pbn.tex
\newif\ifeusipstyle
\newif\ifdohybrid
\dohybridfalse

\documentclass[journal]{IEEEtran}

\usepackage[dvips]{graphicx}
\usepackage{url}
\usepackage{amsfonts}
\usepackage{verbatim}
\usepackage{color}
\usepackage{tablefootnote}
\usepackage{acronym}
\usepackage{booktabs,amsmath,tabularx,multirow,adjustbox}
\acrodef{cnn}[CNN]{convolutional neural network}
\acrodef{dnn}[DNN]{deep neural network}
\acrodef{hmm}[HMM]{hidden Markov model}
\acrodef{pbn}[PBN]{projected belief network}
\acrodef{pbn-da}[PBN-DA]{discriminative alignment of \acl{pbn}}
\acrodef{pdf}[PDF]{probability density function}

\title{Projected Belief Networks With Discriminative
Alignment for Acoustic Event Classification: Rivaling State of the Art CNNs }

\author{\IEEEauthorblockN{Paul M. Baggenstoss}
   \IEEEauthorblockA{Fraunhofer FKIE,
   Fraunhoferstrasse 20\\
   53343 Wachtberg, Germany\\
   Email: p.m.baggenstoss@ieee.org\\}
   \and
   \IEEEauthorblockN{Kevin Wilkinghoff}
   \IEEEauthorblockA{Fraunhofer FKIE\\
   Email: kevin.wilkinghoff@ieee.org\\}
   \and
   \IEEEauthorblockN{Felix Govaers}
   \IEEEauthorblockA{Fraunhofer FKIE\\
   Email: felix.govaers@fkie.fraunhofer.de\\}
   \and
   \IEEEauthorblockN{Frank Kurth}
   \IEEEauthorblockA{Fraunhofer FKIE\\
   Email: frank.kurth@fkie.fraunhofer.de\\}
   \thanks{This work was supported jointly by the Office of Naval Research Global
   and the Defense Advanced Research Projects Agency under Research Grant - N62909-21-1-2024}
}

\begin{document}

\include{macros}
\newtheorem{identity}{Identity}
\newtheorem{hypothesis}{Hypothesis}
\newcommand{\mathtiny}[1]{\mbox{\tiny$#1$}}

\maketitle

\begin{abstract}
The projected belief network (PBN) is a generative stochastic
network with tractable likelihood function 
based on a feed-forward neural network (FFNN). The
generative function operates by ``backing up" through
the FFNN. The PBN is two networks in one,
a FFNN that operates in the forward direction,
and a generative network that operates in the backward direction.
Both networks co-exist based on the same parameter set, have 
their own cost functions, and can 
be separately or jointly trained.  The PBN therefore
has the potential to possess the best qualities of
both discriminative and generative classifiers.
To realize this potential, a separate PBN is trained on
each class, maximizing the generative likelihood function for the given
class, while minimizing the discriminative 
cost for the FFNN against ``all other classes". This technique, 
called {\it discriminative alignment} (PBN-DA), aligns the contours of the
likelihood function to the decision boundaries and attains
vastly improved classifcation performance, rivaling that
of state of the art discriminative networks.
The method may be further improved using a hidden Markov
model (HMM) as a component of the PBN, called PBN-DA-HMM.
This paper provides a comprehensive treatment of
PBN, PBN-DA, and PBN-DA-HMM. In addition, 
the results of two new classification experiments are provided.  
The first experiment uses air-acoustic events,
and the second uses underwater acoustic data
consisting of marine mammal calls.
In both experiments, PBN-DA-HMM attains comparable or
better performance as a state of the art CNN, and
attains a factor of two error reduction when combined with the
CNN.  
\end{abstract}

\section{Introduction}
\subsection{Generative vs. Discriminative Classification}
Classification is probably the most important task in machine learning
because it involves making decisions between several possible ways to understand the data.
%which we call hypotheses or data classes, or simply {\it classes}.  
Until the advent of deep learning, this task was best solved by the human mind
in conjunction with human senses and perception.  There remain two distinct approaches 
to classification: the generative approach, and the discriminative approach
\cite{Goodfellow2016, Vapnik99}.
Both approaches find parallels in human perception and decision-making.

In the generative approach, an attempt is made to characterize each  
class individually by estimating a set of patterns or parameters, then later 
compare the data to each stored characterization.  In statistical terms, 
the probability distribution of the data is estimated for each class hypothesis,
and later the {\it likelihood} of the data is determined with respect
to each learned distribution estimate.  In human perception, the
generative approach corresponds to recognizing something
based on our memory of what the class should look or sound like.

In the discriminative approach, an attempt is made to distinguish between the existing
class hypotheses, an approach that is inherently more efficient
and generally provides performance superior to the generative approach \cite{Vapnik99}.
In human perception, the discriminative approach corresponds
to looking or listening for characteristic
clues that can distinguish between hypotheses.  For example,
if we know that two identical twins differ only in a birthmark, we look 
for the presence of this hard-to see characteristic.

The generative approach is inherently inefficient because
it is {\it blind} to the other class hypotheses, so
does not know in advance which subtle clues it needs, and must 
extract a complete detailed characterization of each class hypothesis. 
As the data dimension increases, the amount of information required 
to adequately characterize the data distribution rises exponentially, 
outstripping what is available in the training data. This is the 
so-called curse of dimensionality \cite{Vapnik99}.  The generative approach 
becomes even more difficult in noisy
scenarios because generative models cannot distinguish between the target
signal and noise.

Despite these problems, generative classifiers are very useful,
especially in open-set classification problems
where they can detect out-of-set events.  In addition, having a 
``second opinion" from a generative approach can result in 
improved performance for a combined generative/discriminative approach.

\subsection{Attaining the best properties of both approaches}
When combining classifiers, such as by additively combining the
classifer statistics (i.e. ensembling), it is important that the
two methods (a) have comparable performance and (b)
are based on independent  views of the data.
This is why it is important to create better-performing
 generative classifiers to combine with purely discriminative ones.
Therefore, it would be desireable if generative classifiers
could be constructed to utilize discriminative information,
analogous to how humans do, and thereby achieve improved performance.
Later, in the experiments, we will combine PBN with 
a discriminative CNN, but this is not what we mean by a combined approach.
This is because the two methods are not able to benefit from each other
during training.  Using PBN, it is possible
to apply both the generative and discriminative approaches
simultaneously, that is to create a generative model with discriminative capability.  
Even if this approach does not improve upon state of the art discriminative classifiers,
it could result in a classifier that is better for combination with them.  

In human perception, we use a combined generative and discriminative  approach
if we give discriminative clues priority as we collect clues for the generative approach. 
To use the example of the identical twins above, we would
add the presence of a birthmark to the list of features that describe
each class individually.  So, conceptually,  a combined
approach is not hard to understand.  Does there
exist a parallel idea in machine learning, in which a given feature 
extractor (i.e. a given network) can be both generative and discriminative?

Many methods have been proposed that either incorporate both generative
and discriminative ideas or achieve the goal of joint
generative and discriminative training 
\cite{ZhengJoint2019,GORDON2020107156,TuGenDisc07,RainaSNM03,Liu2020HybridDT,NIPS1998_db191505}.
But these approaches concentrate on training methods, and do 
not treat the problem at the fundamental level that we are proposing.

\subsection{Discriminative and Generative functions in a single network}
\label{dg1sec}
To have both discriminative and generative functions in a single network,
the network needs to operate in both the forward and backward
directions.  We will call this a two-directional network,
so as not to confuse the idea with bi-directional networks\footnote{Bi-directional networks
have some layers that operate forward
in time and some layers that operate backward in time. } \cite{SchusterBidirectional} because it can operate
in the forward direction (i.e. as a feed-forward classifier network), 
but also in the backward direction (as a generative network), using the same 
network architecture and parameter set.
We seek the following properties of a two-directional network:
\benum
\item
The desired network topology must be dimension-reducing in order to serve
also as a classifier.  
\item
The generative function should be complete, which means it should provide
a tractable likelikood function (LF) (i.e. log of the
probability density function) of the input data, and should be able to 
synthesize samples corresponding to this LF.
\item
To be a truly dimension-reducing two-directional network, we propose that the reverse path should
constitute a {\it right inverse} operation. If the forward path is
written $\bfz=T(\bfx)$, where $\bfx\in \mathbb{R}^N$, $\bfz\in \mathbb{R}^M$, $N>M$, 
and the right inverse is written  $\bfx=T^{-1}(\bfz)$, then we must have $T\left( T^{-1}(\bfz)\right) = \bfz$. 
This is important because or stated goal is that the discriminative
(forward path) and generative (backward path) should learn jointly.
The right-inverse property insures a tight connection between the two directions.
In fact, it should be clear that the right-
inverse property can only be achieved 
by a single network, i.e. not by separate networks that each operate in 
one direction, because the reverse path is the inverse of the forward path.  

\eenum

There are a number of existing machine 
learning approaches that we can consider.

Although they are not right-inverse networks, we discuss two related
approaches: auto-encoders and stacked restricted Boltzmann machines (RBMs).
The RBM consists of two back-to-back stochastic perceptron layers
with tied weights (the backward weight matrix
is the transpose of the forward weight matrix) \cite{HintonDeep06}.
Although the reverse path of the RBM is not a right inverse, 
it uses the same weight matrix and  
is a generative network based on an elegant theory, so
is worth looking at.
The input data (visible data) and the hidden variables
are jointly distributed according to the Gibbs distribution.
The RBM is trained using an approximate maximum likelihood
approach called  contrastive divergence (CD) \cite{WellingHinton04}.
Once trained, either direction of the RBM can be converted into a perceptron
layer by replacing the stochastic perceptrons with 
deterministic activation functions equal to their expected value.
Multiple deterministic RBM perceptron layers can then be cascaded together 
(in both directions) to obtain a stacked RBM.  The stacked RBM has 
also a statistical interpretation
and can be trained by a multi-layer version of CD
called up-down algorithm \cite{HintonDeep06}.

A stacked RBM is a multi-layer two-directional network
without right-inverse, and is identical to an
auto-encoder with tied analysis and synthesis weights, but
trained differently.  Note that it is not uncommon to tie the analysis and synthesis weight matrices 
of an auto-encoder (i.e. so that their corresponding
linear operations are transposed versions of each other), which
is possible both for dense and convolutional layers \cite{LiTied2019}.
Auto-encoders are in wide-spread use because it is inherently useful
to find a low-dimensional output feature set (called the encoding or {\it bottleneck} features)
with minimum dimension that can be used to reconstruct the input data.

In addition to lacking a right-inverse, these networks are also not
complete generative models, because they lack a tractable LF.
Even the stacked RBM, which has a well-defined statistical model \cite{HintonDeep06},
cannot be tractably marginalized  to obtain the likelihood function of the
input data.  
%Although the reconstruction path in these approaches does not constitute a right-inverse,
%they deserve attention because the forward network can be 
%jointly trained as a classifier, if it is extended past the bottleneck
%layer and ends in a classification output layer.  Then, a combined cost function
%can be used to train the network for both tasks, reconstructing the 
%input data, and acting as a classifier \cite{GovaersBagFusion}.  
%As we will mention later in Section \ref{pbn_init_sec},
%the network obtained this way can be used as the initial parameters
%of a PBN, which is a two-directional
%generative network, having tractable LF and being the right-inverse of the forward path.
A variational auto-encoder (VAE) is a type of auto-encoder
\cite{odaibo2019tutorial,doersch2021tutorial,BetaVAE},
but the LF must be approximated and it is not a right-inverse network.

Normalizing flows (NF) \cite{kobyzev2021normalizing} provide a way of defining a probability distribution
by finding a 1:1 dimension-preserving transformation that maps the input data to 
a well-defined distribution such as to independent identically-distributed ({\it iid})
uniform or normal random variables. The distribution of the 
input data is found using the change of variables theorem, i.e. 
using the determinant of the transformation's Jacobian matrix. The forward path extracts the 
{\it iid} random variables, and the reverse path creates 
synthetic input data. The problem with NF is that the forward path is not
dimension-reducing, so cannot be used as a discriminative classifier.
That is unfortunate because NF is truly two-directional
and has a well-defined likelihood function.
In order to get around the requirement
of dimension-preserving transformations, a way to combine NF with variational autoencoders (VAEs)
called ``SurVAE" has been proposed, \cite{Nielsen_SurVAE_NEURIPS2020}.
However, for the case of dimension-reducing deterministic
transformations (surjections), 
SurVAE is the same as and is a re-invention of PDF projection 
\cite{BagSurVAE-PBN-Arxiv}, which is the basis of
 our proposed approach and will be reviewed in the next section.

\subsection{What's New in This Paper}
The approaches presented in this paper can all be found in
various forms in existing conference and journal publications
\cite{BagPDFProj,BagMaxEnt2018,Bag_info,BagKayInfo2022,BagUMS,BagEusipco2014ModelMix,BagAESModelMix,
BagIcasspPBN,BagPBNHidim, BagPBNEUSIPCO2019,BagSPL2021}.
However, there is no comprehensive and cohesive treatment
to make the idea set accessible to readers. This paper seeks to fill this void.
The experimental results are also new, as well as the
way the concepts are introduced, specifically
in the context of a two-directional network.

\section{Review of PDF Projection}
\label{pdfprojsec}
\subsection{Definition of PDF Projection}
Before we introduce the projected belief network (PBN),
we need to introduce and review PDF projection, the principle upon which it is based.
Consider a fixed and differentiable dimension-reducing transformation
$\bfz=T(\bfx)$, where $\bfx \in \mathbb{X}\subseteq \mathbb{R}^N$, and $\bfz \in \mathbb{R}^M$, where $M<N$, and whose $N\times M$ matrix of partial derivatives is of full rank.
%Subject to mild constraints \cite{BagPDFProj,BagMaxEnt2018,Bag_info}, and assuming 
Then, assuming a known or assumed
feature distribution $g(\bfz)$, one can construct a probability density function (PDF)
on the input data with support $\mathbb{X}$ given by
\beq
G(\bfx) = \frac{p_{0,x}(\bfx)}{p_{0,x}(\bfz)} g(\bfz),
\label{ppt0}
\eeq
where $p_{0,x}(\bfx)$ is a prior distribution 
with support on $\mathbb{X}$ and $p_{0,x}(\bfz)$
is its mapping to $\mathbb{R}^M$ through $T(\bfx)$. 
In our simplified
notation, the argument of the distribution defines its
range of support, and the variable in the subscript defines the original
range where the distribution was defined.
Therefore, $p_{0,x}(\bfz)$ is a distribution over the range of $\bfz$, but
is a mapping of a distribution that was defined on $\mathbb{X}$.
Note that $G(\bfx)$ is seen as a function only of $\bfx$ since
$\bfz$ is deterministically determined from $\bfx$ using $\bfz=T(\bfx)$.
In fact, it can be shown \cite{BagPDFProj,Bag_info} that 
$G(\bfx)$ is a PDF (integrates to 1) and is a member of the set of PDFs
that map to $g(\bfz)$ through $T(\bfx)$.
If $p_{0,x}(\bfx)$ is selected for maximum entropy, then $G(\bfx)$ is unique for a given
transformation, data range $\mathbb{X}$, and a given $g(\bfz)$ (where ``g" represents
the ``given" feature distribution).  
One can think of $G(\bfx)$ as a constructed distribution based
on a set of parameters (i.e. the parameters of $T(\bfx)$, which could be a neural network for example).
Then, using the method of maximum likelihood, the PDF of the input data can be estimated by training the
parameters of the transformation to maximize the mean of
$\log G(\bfx)$ over a set of training data. This in fact results in a transformation
that extracts sufficient statistics and maximizes information \cite{BagKayInfo2022}
(see Section \ref{infosec}).
We say that $G(\bfx)$ is the ``projection" of $g(\bfz)$ back to the input data
range $\mathbb{X}$, i.e. a {\it back-projection}.
We call the term $J(\bfx)\defined {p_{0,x}(\bfx) \over p_{0,x}(\bfz)}$ the ``J-function"
because in the special case of  dimension-preserving transformations
(i.e. normalizing flows \cite{kobyzev2021normalizing}),
$J(\bfx)$ is the determinant of the Jacobian of $T(\bfx)$.  
The log of the J-function is analogous to the ``likelihood contribution"
found in \cite{Nielsen_SurVAE_NEURIPS2020}.

\subsection{Data Generation}
To generate data from $G(\bfx)$ in (\ref{ppt0}), one
draws a sample $\bfz$ from $g(\bfz)$, then draws a sample $\bfx$ from the 
level set ${\cal L}(\bfz)$ defined by
\beq
  {\cal L}(\bfz) = \{ \bfx \in \mathbb{X} | T(\bfx) = \bfz \},
  \label{levsetdef}
\eeq
and weighted by the prior distribution $p_{0,x}(\bfx)$.
In other words, we sample from $p_{0,x}(\bfx)$ restricted 
and nomalized on ${\cal L}(\bfz)$.

Depending on the transformation $T$ and prior distribution $p_{0,x}(\bfx)$,
sampling may be straight forward, or may require Markov Chain Monte Carlo (MCMC) methods.  
PDF projection sampling for specific examples of transformations
are treated in \cite{Bag_info, BagUMS}.

%In the class of stochastic layered generative networks, with 
%a dimension-increasing generation process, the PBN is unique
%in that it has a tractable LF.

\subsection{Chain Rule}
In order to implement (\ref{ppt0}),
it is necessary to know $p_{0,x}(\bfz)$.
For complex non-linear transformations, it may be
impossible to find a tractable form for $p_{0,x}(\bfz)$.
However, using the chain-rule, it is possible to 
solve the problem by breaking the transformation into 
a cascade of simpler transformations and applying the 
PDF projection idea recursively
to stages of a transformation.
Consider a cascade of two transformations, $\bfy=T_1(\bfx)$, and $\bfz=T_2(\bfy)$.
Then, applying (\ref{ppt0}) recursively,
\beq
G(\bfx) = \frac{p_{0,x}(\bfx)}{p_{0,x}(\bfy)} \frac{p_{0,y}(\bfy)}{p_{0,y}(\bfz)} g(\bfz),
\label{ppt1}
\eeq
which can be extended to any number of stages.
To compute $\log G(\bfx)$, one just accumulates the log-J function of the transformations.
Data generation is also cascaded, and is initiated by drawing a sample $\bfz$ from $g(\bfz)$.
% we talked about this already?
%Note that $p_{0,x}(\bfy)$ and $p_{0,y}(\bfy)$ are two different distributions
%on the range of $\bfy$. While $p_{0,y}(\bfy)$ is originally defined on
%the range of $\bfy$, $p_{0,x}(\bfy)$ can be written $T_1[p_{0,x}(\bfx)].$

\subsection{Multiple features}
\label{pdfpmultif}
In recent years, end-to-end deep learning approaches
are slowly making feature extraction obsolete.
But, generative classification approaches are
rarely used on unprocessed data. Some form of feature extraction, such as
conversion to time-frequency spectral features, is carried
out before application of generative models because this tends to reduce dimensionality
and remove nuisance information such as phase.
However, data can sometimes have very diverse character
and just one feature extraction approach may not suffice.

But, the problem with using multiple feature extraction
approaches in generative models is that 
the definition of ``input data" is not clear when there
are several feature sets.  This problem is solved by PDF projection because
all feature extraction transformations 
can project back to the same input data space.

There are a number of approaches to using PDF projection for this purpose.
One possible approach is to assign an individually optimized feature set to each data class.
But, using one feature extraction for {\it each} data class can be
problematic.  For example, the existence of background noise in the data could make a different feature
extractor more ``suitable", increasing the likelihood
values for that feature/class combination and
 cause classification errors.  Therefore,
it is safer and more effective to
use the same set of feature extraction approaches for all classes.
Using PDF projection, the likelihood function computed for each feature
after back-projecting to the raw data can be seen as a component of a mixture distribution
\cite{BagEusipco2014ModelMix,BagAESModelMix}. This brings much more information to bear on the
problem without increasing the dimensionality of the generative models.

To add mathematical detail, consider $L$ feature extraction chains, resulting in features $\bfz_1, \ldots \bfz_L$, 
with feature probability density functions $g_1(\bfz_1), \ldots g_L(\bfz_2)$. 
Applying (\ref{ppt0}) separately to each model, we arrive
at the projected PDFs $G_1(\bfx), \ldots G_L(\bfx)$. 
The mixture distribution can be created as follows
$$G(\bfx) =  \sum_{i=1}^L  \;\alpha_i \;G_i(\bfx).$$
Such a mixture model is then trained separately on each class, with the weights $\{\alpha_i\}$ also considered
as parameters to estimate.

However, a problem that arises when using multiple feature sets is caused
by the widely-used technique of overlapped, windowed segmentation
of time-series.  While it is reasonbable to compare likelihood values for
features extracted from the same input data, 
it is not clear about how to resolve the differences between two feature extractors operating on
essentially different raw data.  When applying different overlapped segmentation
prior to feature extraction, the definition of ``raw data" changes with the segmentation
window-size.  A solution to this problem is provided by what we call Hanning-3 segmentation, \cite{BagHanning},
however this it out of scope for this paper.

\subsection{Information Maximization by PDF Projection}
\label{infosec}
Dimension-reduction is an important function in 
neural networks and machine learning with many
end-goals including classification, auto-encoding, and feature embedding.  
When the end-goal is not known in advance, one can
try to maximize the information at the output.  
A widely-used criteria for maximizing information transfer
is the InfoMax principle, which seeks to maximize the mutual information (MI) between
the input data $\bfx$ and output data $\bfz$
\cite{ZhuRadar,NadalInfo,lin2006conditional,DecoDragan}.
The mutual information can be decomposed
as $$I(\bfx,\bfz)=H(\bfz)-H(\bfz|\bfx),$$ where
$H(\bfz)$ and $H(\bfz|\bfx)$ are the entropy and conditional entropy
\cite{NadalInfo,KayInfo}.  For noiseless deterministic transformations,
$\bfz$ is known perfectly given $\bfx$, so $H(\bfz|\bfx)=0$ \cite{NadalInfo,KayInfo}.
Thus, the task comes down to maximizing the entropy of 
the network output $\bfz$ \cite{NadalInfo}.
But this is not very satisfactory as an information metric because
$H(\bfz)$ can be made as large as possible simply by 
scaling or applying 1:1 transformations  which do not change the
information content.  To arrive at meaningful results with InfoMax,
one has to impose an arbitrary constraint, such as variance or data range.
A more satisfactory criterion was proposed in \cite{BagKayInfo2022}.
The proposed approach is to quantify the separability
of the input data distribution $p(\bfx)$ from a maximum entropy (MaxEnt) reference
distribution $p_0(\bfx)$ through the Kullback-Leibler divergence
(KLD), written $D( p \| p_0)$, where  $D( p \| q)$ is the KLD :
\beq
D( p \| q) \defined \mathbb{E}_p \left\{ \log \frac{p(\bfx)}{q(\bfx)}\right\}.
\label{kld}
\eeq
A MaxEnt distribution $p_0(\bfx)$ is bland and featureless, so it makes sense
that the data distribution $p(\bfx)$ stands out relative to it.
The idea is that for maximum information flow, 
the separability of the two distributions should not be reduced
by the network, or in other words, we try to maximize
the divergence at the output, which is bounded by the 
divergence at the input. It is shown in \cite{BagKayInfo2022},
that this same goal is achieved by PDF projection when one
estimates the distribution the input data by
fitting $G(\bfx)$ to $p(\bfx)$ using maximum likelihood
and the available training data: 
\beq
\max_T \left\{ \sum_i \log G(\bfx_i;T,g) \right\},
\label{ifxf1}
\eeq
where we have added $T$ and $g$ as arguments to $G(\bfx)$
to make clear that (\ref{ppt0}) depends on transformation $T$ and given feature density $g$.
It is interesting to note that as (\ref{ifxf1}) is maximized
over $T$, the true feature distribution is in fact driven toward $g(\bfz)$.
A side-benefit of this is that if $g(\bfz)$ is some canonical
distribution of ({\it iid})
output variables, then the true feature distribution is driven toward
$g(\bfz)$, i.e. $p(\bfz)\longrightarrow g(\bfz)$.
Therefore, (\ref{ifxf1}) achieves three goals at the same
time :
\benum
\item Maximization of information flow. 
\item PDF estimation, i.e. $G(\bfx_i;T,g) \longrightarrow p(\bfx)$.
\item Generation of information-bearing statistics $\bfz$ with 
arbitrary desired distribution as $p(\bfz)\longrightarrow g(\bfz)$.
\eenum

\section{Projected Belief Network (PBN)}

\subsection{Definition of PBN}
When PDF projection is applied to a feed-forward neural network (FFNN) layer-by-layer, this results in the
projected belief network (PBN) \cite{BagPBN}.
The PBN is a true two-directional network because the reverse path (sampling) is carried out 
backwards through the FFNN and is a right-inverse.
%Although the hidden variables $\bfz$ are random and jointly
%distributed with $\bfx$, the marginalization integration 
%to obtain the likelihood function (LF) $p(\bfx)$  occurs on a
%manifold of zero volume (see ${\cal M}(\bfz)$ below), with the result
%that the PBN has a tractable likelihood function (LF).
%The PBN is stochastically sampled by ``backing up" through the FFNN. 
%
%As an aside, a PBN can also be sampled deterministically by choosing the conditional mean \cite{BagIcasspPBN},
%resulting in the deterministic PBN (D-PBN).  By passing forward through the FFNN, then backing
%up using D-PBN, a type of auto-encoder is formed \cite{BagIcasspPBN,BagPBNHidim, BagPBNEUSIPCO2019, BagPBN}.
%

\subsection{Mathematical Details of a PBN Layer}
\label{mandist}
We concentrate on a single PBN layer, which corresponds to
one perceptron layer in a neural network.
To illustrate the sampling process in one layer, let $\bfy\in\mathbb{R}^M$ be the
hidden variable at the output of a given layer of a FFNN,
and let ${\bfx\in\mathbb{R}^N}$ be the layer input, where
$N>M$.  Let $\bfy=\lambda\left({\bf b}+{\bf W}^\prime \bfx\right),$
where ${\bf W},{\bf b}$ are the layer weight matrix and bias,
and $\lambda(\;)$ is a strictly monotonic increasing (SMI) element-wise activation function.
We seek to sample $\bfx$ given $\bfy$.  Because the SMI activation function and bias 
are invertible, it is equivalent to sample $\bfx$ given $\bfz$, where
$\bfz={\bf W}^\prime \bfx.$  This is done by drawing $\bfx$ randomly from  the set 
of samples that map to $\bfz$, weighted by prior distribution $p_{0,x}(\bfx)$.
More precisely, $\bfx$ is drawn from level set (\ref{levsetdef}), which
in the case of a linear transformation is the set (an affine subset and
a manifold) 
\beq
  {\cal M}(\bfz) = \{ \bfx \in \mathbb{X} | {\bf W}^\prime \bfx = \bfz\},
  \label{mandef}
\eeq
with prior density $p_{0,x}(\bfx)$. This prior is 
selected according to the principle of maximum entropy (MaxEnt) \cite{Jaynes57},
so it has the highest possible entropy subject to any constraints.
These constraints include the range of the variable $\bfx$, denoted by $\mathbb{X}$, determined
by which activation function was used in the up-stream layer,
and some moment constraints that may be necessary when $\mathbb{X}$ is unbounded.
By definition, the drawn sample will be a right-inverse 
(i.e. preimage) of $\bfz$.

Note that when discussing PBNs, we use the variable $\bfx$ to denote 
either a layer input or a network input, depending on the context.
The variable $\bfz$ always refers to the output of the linear transformation,
and the variable $\bfy$ always refers to the output of an activation function,

We consider three canonical data ranges that are common in machine mearning, 
$\mathbb{R}^N$ : ${x_i\in (-\infty , \infty), \forall i}$, $\mathbb{P}^N$ : ${x_i\in [0 , \infty), \forall i}$, and
$\mathbb{U}^N$ : ${x_i\in [0  , 1], \forall i}$. In Table \ref{tab1a},
we list four useful combinations of data range $\mathbb{X}$ and MaxEnt 
prior $p_{0,x}(\bfx)$.  Note that when a data range is defined that 
is unbounded, the entropy of a distribution with support on
that range can go to infinity.  Therefore, to find a MaxEnt distribution,
it is necessary to place constraints on the moments, either variance or mean.
This is why the MaxEnt distributions are defined either
with a fixed variance (Gaussian or truncated Gaussian) or
fixed mean (exponential).
\begin{table}
    \caption{MaxEnt priors and activation functions as a function of input data range.
        TG=``Trunc. Gauss.". TED=``Trunc. Expon. Distr".
        ${\cal N}\left(x\right) \defined \frac{e^{-x^2/2}}{\sqrt{2\pi}}$ and $\Phi\left( x\right)  \defined \int_{u=-\infty}^x {\cal N}\left(u\right) {\rm d}u.$
%Note that the ``exponential" distribution is more 
%       precisely the chi-squared distribution with 2 degrees of freedom, having mean 2.
    }
    \begin{center}
        \begin{tabular}{|l|l|l|l|l|}
            \hline
     $\mathbb{X}$ & \multicolumn{2}{|c|}{MaxEnt Prior} & \multicolumn{2}{|c|}{Activation} \\
            \hline
              &  $p_{0,x}(\bfx)$  &  Name & $\lambda(\alpha)$  & Name \\
            \hline
            $\mathbb{R}^N$   & $\prod_{i=1}^N {\cal N}(x_i)$  & Gaussian & $\alpha$  & linear\\
            \hline
            $\mathbb{P}^N$   & $\prod_{i=1}^N 2 {\cal N}(x_i)$ & Trunc. Gauss. & $\alpha + \frac{{\cal N}(\alpha)}{\Phi(\alpha)}$  & TG \\
            \hline
            $\mathbb{P}^N$   & $\prod_{i=1}^N  \frac{1}{2} e^{-x_i/2}$ & Expon. & $\frac{2}{1-2\alpha}$, $\;\;\alpha<.5$& Expon. \\
            \hline
            $\mathbb{U}^N$  & $\;\;\;\;\;\;$ 1 $\;\;\;\;\;\;$ &   Uniform & $\frac{e^{\alpha}}{e^{\alpha} - 1}-\frac{1}{\alpha}$ & TED \\
            \hline
        \end{tabular}
    \end{center}
    \label{tab1a}
\end{table}
Additional details can be found in \cite{BagIcasspPBN,BagPBNHidim}.

The PDF of the input data of a PBN layer is given by the projected PDF (\ref{ppt0}),
which can be seen as an application of Bayes rule \cite{Bag_info}  $G(\bfx)=p_m(\bfx|\bfz) g(\bfz)$,
where  $p_m(\bfx|\bfz)$ is the {\it a posteriori} distribution,
also equal to the J-function, i.e. 
\beq
p_m(\bfx|\bfz) =  \frac{p_{0,x}(\bfx)}{p_{0,x}(\bfz)} = J(\bfx).
\label{jpostdef}
\eeq
Note that $p_m(\bfx|\bfz)$ is a manifold distribution with support only on a set of zero volume, the manifold (\ref{mandef}).
To see this, we integrate the J-function on the manifold for fixed $\bfz$:
\beq
\int_{\bfx\in{\cal M}(\bfz)} \; \frac{p_{0,x}(\bfx)}{p_{0,x}(\bfz)}  \; {\rm d} \bfx = \frac{1}{p_{0,x}(\bfz)}  \; \int_{\bfx\in{\cal M}(\bfz)} \; p_{0,x}(\bfx) \; {\rm d} \bfx .
\label{jfint}
\eeq
The denominator of (\ref{jfint}) is independent of $\bfx$ on the manifold (\ref{mandef}), and we have the identity
$$\int_{\bfx\in{\cal M}(\bfz)} \; p_{0,x}(\bfx) \; {\rm d} \bfx  = p_{0,x}(\bfz).$$
Therefore, $p_m(\bfx|\bfz)$ integrates to 1 on the manifold.

\subsection{Sampling a PBN }
\label{sampdefsec}
We now summarize the stochastic sampling of a multi-layer
PBN. We assume that the last layer of the network does not use an activation function (for theoretical purposes we can ignore the
activation functions because they are assumed to be SMI and can be inverted).
Let the network output be denoted by $\bfz$.
Sampling proceeds as follows starting at the last (output) layer:
\benum
\item Set the network layer counter to the last layer. 
\item We start by drawing a sample $\bfz$ from the network output feature distribution $g(\bfz)$.
\item
    Draw a sample from the manifold distribution $p_m(\bfx|\bfz)$,
i.e. draw a sample fom ${\cal M}(\bfz)$, weighted by the
prior $p_{0,x}(\bfx)$.  How this is done depends on the
 prior distribution $p_{0,x}(\bfx)$, which is selected from Table \ref{tab1a}.
 For the Gaussian case, drawing from $p_m(\bfx|\bfz)$ is straight-forward,
  but for all the other cases, it is necessary to find a starting
solution, any member of ${\cal M}(\bfz)$, then use
a form of MCMC to draw a sample from
set ${\cal M}(\bfz)$ weighted by prior $p_{0,x}(\bfx)$ \cite{BagUMS}.
As an initial solution, one can either use a linear programming solver,
but a better starting point is the conditional mean
$\bar{\bfx}_z$ as given later in Section \ref{asysec}.
If no solution exists, sampling has failed, return to step 1.
%\item Assume that the current layer's weight matrix is given by ${\bf W}$.
%  Use some linear programming solver to obtain an initial right-inverse solution, i.e. 
%any member of (\ref{mandef}).  Alternatively, we can solve for the manifold centroid, $\bar{\bfx}_z$ as
%given later in equation (\ref{meanzh}).  If no solution exists, sampling has failed, return to step 1.
%By counting the number of failed samples, the network sampling efficiency can be estimated.
%\item Starting with the initial solution $\bfx$ from the previous step, use 
%a form of Monte Carlo Markov chain (MCMC) to draw a sample from
%set (\ref{mandef}) weighted by prior $p_{0,x}(\bfx)$.
%This is the same as drawing a sample randomly from the 
%{\it a posteriori} distribution $p(\bfx|\bfz)$, 
%\beq
%p(\bfx|\bfz) = \frac{p_{0,x}(\bfx)}{p_{0,x}(\bfz)},
%\label{cndxz}
%\eeq
%which has support only on (\ref{mandef}), as explained in Section \ref{mandist}.
%The reference distribution $p_{0,x}(\bfx)$ is selected from Table \ref{tab1a}.
%Depending on $p_{0,x}(\bfx)$, $p_{0,x}(\bfz)$ is either known exactly, or its moment generating function (MGF)
%is known exactly and needs to be inverted (See \cite{BagSPL2021}).
%Sampling from $p(\bfx|\bfz)$ can be done by Monte Carlo Markov chain (MCMC)
%by finding an initial solution $\bfx$, then moving
%along orthogonal directions in the column space of ${\bf W}$, making sure
%to weight according to $p_{0,x}(\bfx)$. Sampling for some cases is detailed in
%\cite{Bag_info}. 
\item
  If this current layer is the first network layer, then sampling is finished,
and $\bfx$ is a sample of the input data. 
 If this is not the first layer, then consider $\bfx$ to be the
output $\bfy$ of the prevous layer's activation function. 
To proceed, work backward by inverting the activation function and bias 
of the previous layer to obtain the previous layer's linear transformation output variable $\bfz$.
Set layer counter to the previous layer, and proceed with step 3.
\eenum

\subsection{Asymptotic Form of PBN}
\label{asysec}
As elegant as (\ref{jpostdef}) may be, it is difficult to work with. 
For example, integrating $p_m(\bfx|\bfz)$ to get the
conditional mean, 
\beq
\bar{\bfx}_z = \mathbb{E}_{x|z} = \int_{\bfx\in{\cal M}(\bfz)} \; \bfx \; p_m(\bfx|\bfz) \;  {\rm d} \bfx
\label{meanzh1}
\eeq
only has a closed form in case of a Gaussian prior.
This is a result of the fact that the denominator $p_{0,x}(\bfz)$
is otherwise intractable.

Luckily, an ``almost exact" approximation to $p_{0,x}(\bfz)$
can be found, resulting in a simple functional form
for $p_m(\bfx|\bfz)$ for which the conditional mean can be
found in closed form.  
Steven Kay \cite{BagNutKay2000} noted that
in cases like this (linear combinations of non-Gaussian
random variables), the moment generating function 
(MGF) of $p_{0,x}(\bfz)$ is in fact known exactly, and
can be inverted to find $p_{0,x}(\bfz)$ using
the saddle point approximation (SPA).
The word ``approximation" here is misleading because the inversion of the MGF,
is ``almost" exact.  Although it is only a first-order approximation, the SPA
integral has two important advantages: the integrand is known exactly, 
and the integral is carried out at the
saddle point. As $N$ becomes large, the 
integrand, as a result of the central limit
theorem, takes an approximately Gaussian shape
at the saddle point, which
is exploited by the SPA to result in a solution for $p_{0,x}(\bfz)$
that is exact {\it for all practical purposes}.

Note that ideally, $p_m(\bfx|\bfz)$ has support only on
${\cal M}(\bfz)$, but the ``almost exact" functional form 
for $p_m(\bfx|\bfz)$ , called the {\it surrogate density}
 \cite{BagPBN,BagIcasspPBN}, is a proper distribution in
its own right, with support in all of $\mathbb{X}$.  However,
as $N$ becomes large, the probability mass 
of the surrogate density converges around the manifold ${\cal M}(\bfz)$
(see Section VII-A in \cite{BagUMS})
and the resulting closed-form expression for $\bar{\bfx}_z$
converges quickly to the true value \cite{BagSPL2021}.

The surrogate density as well as the prior $p_{0,x}(\bfx)$
for all cases of $\mathbb{X}$ in Table \ref{tab1a}
can be represented by a single exponential-form density as follows
\beq
p_s(\bfx;\balpha,\alpha_0,\beta)=\prod_{i=1}^N \; p_e(x_i;\alpha_i,\alpha_0,\beta),
\label{padef}
\eeq
where $p_e(x;\alpha,\alpha_0,\beta)$ is a univariate distribution of the exponential class
\beq
\log p_e(x;\alpha,\alpha_0,\beta) = (\alpha_0+\alpha) x + \beta x^2 + \log Z(\alpha,\alpha_0,\beta),
\label{expclassdef}
\eeq
where $\balpha=\{\alpha_1, \ldots \alpha_M\}$,
and $\log Z(\alpha,\alpha_0,\beta)$ is the the normalization factor 
necessary such that $p_e(x;\alpha,\alpha_0,\beta)$ integrates to 1.
We also define the corresponding theoretical activation function using the expected value of $p_e(x;\alpha,\alpha_0,\beta)$
$$\lambda(\alpha,\alpha_0,\beta)= \int_x \; x\; p_e(x;\alpha,\alpha_0,\beta) {\rm d}x.$$ 
In Table \ref{tab1v}, we provide $p_e(x;\alpha,\alpha_0,\beta)$ for various
choices of $\mathbb{X}$, $\alpha_0$, and $\beta$. 
The prior is obtained for $\balpha={\bf 0}$
$$p_{0,x}(\bfx) = p_e(x;{\bf 0},\alpha_0,\beta).$$

%For simplicity, we drop $\alpha_0$, $\beta$ from the notation because it is fixed by the choice of $p_{0,x}(\bfx)$.

\begin{table}[htb!]
\begin{center}
\caption{Extension of Table \ref{tab1a} showing $\alpha_0$, $\beta$, and $p_e(x;\alpha,\alpha_0,\beta)$ for all cases.  }
\begin{tabular}{|c|c|c|c|c|}
\hline
$\mathbb{X}$ & $\alpha_0$ & $\beta$ & $p_e(x;\alpha,\alpha_0,\beta)$  & Name\\
\hline
$\mathbb{R}^N$ & 0 & -.5 & ${\cal N}(x-\alpha)$ & Gaussian.\\
\hline
$\mathbb{P}^N$ & 0 & -.5 & $2 {\cal N}(x-\alpha)$ & Trunc. Gauss. (TG)\\
\hline
 $\mathbb{P}^N$ & -.5 & 0 & $\frac{(1-2\alpha)}{2} e^{-\frac{(1-2\alpha)}{2}  x}$ & Expon.\\
\hline
 $\mathbb{U}^N$ & 0 & 0 & $\left(\frac{\alpha}{e^{\alpha} - 1}\right)  \; e^{\alpha x}$ & Trunc. Expon. (TED)\\
\hline
\end{tabular}
\label{tab1v}
\end{center}
\end{table}

By comparing Tabes \ref{tab1a} and \ref{tab1v},
it can be seen that the MaxEnt priors and activation functions
given in Table \ref{tab1a} are special cases of $p_s(\bfx;\balpha,\alpha_0,\beta)$
where $\balpha={\bf 0}$, and the activation functions are the expected value as a function of $\alpha$.  More precisely,
\beq
p_{0,x}(\bfx)=p_s(\bfx;{\bf 0},\alpha_0,\beta), 
\label{p0def}
\eeq
\beq
\lambda(\alpha)=\int_x \; x \; p_e(x;\alpha,\alpha_0,\beta) {\rm d}x,
\label{l0def}
\eeq
where $\alpha_0$ and $\beta$ are provided in Table \ref{tab1v}.

We restate the following theorem, first published in \cite{BagIcasspPBN}:
\begin{theorem}
Let prior $p_{0,x}(\bfx)$ be written as (\ref{p0def}) , having mean $\lambda({\bf 0})$
as given in (\ref{l0def}).  Then, the surrogate density for $p_m(\bfx|\bfz)$ in (\ref{jpostdef}) is 
 $p_s(\bfx;{\bf W} \bfh_z, \alpha_0,\beta)$, where $\bfh_z$ is value of $\bfh$ that solves
\beq
{\bf W}^\prime \lambda\left( {\bf W} \bfh\right) = \bfz.
\label{tm1}
\eeq
Then, as $N$ becomes large, $p_s(\bfx;{\bf W} \bfh_z, \alpha_0,\beta) \rightarrow p_m(\bfx|\bfz)$.
%Interestingly, using the surrogate density in place of $p_m(\bfx|\bfz)$
%is the same as using the saddle point approximation to replace
%$p_{0,x}(\bfz)$ in (\ref{jpostdef}) \cite{BagIcasspPBN}.  

Furthermore, the mean of the surrogate density is asymptotically
(for large $N$) equal to the mean of $p_m(\bfx|\bfz)$ and is given by
\beq
  \bar{\bfx}_z = \lambda({\bf W} \bfh_z).
  \label{meanzh}
\eeq
\label{thm1}
\end{theorem}
For an outline of the proof, see \cite{BagIcasspPBN}.

In summary, the surrogate density converges to the posterior $p_m(\bfx|\bfz)$,
and so the mean of the surrogate density approaches the mean of $p_m(\bfx|\bfz)$.
This convergence occurs quickly as a function of $N$, as has been demonstrated
in certain cases (see Fig. 8 in \cite{BagUMS}).
It has been shown that the error of the likelihood calculations
when doing this are negligible \cite{BagPBN,BagSPL2021}.
The surrogate density mean $\bar{\bfx}_z$ given by (\ref{meanzh}) enjoys numerous properties.
As conditional mean estimator, it has many well-known optimal properties \cite{KayEst}.
Another special case of (\ref{meanzh})  corresponds to autoregressive spectral estimation,
which can be generalized for any linear function of the spectrum, such as
MaxEnt inversion of MEL band features \cite{BagUMS}.  A special case of (\ref{meanzh}) is mathematically the same as
classical maximum entropy image reconstruction \cite{Wernecke77,Wei87}.
It can also be shown that $\bfh_z$ is the maximum likelihood estimate of  $\bfh$ under the
likelihood function $p(\bfx;\bfh) = p_s(\bfx;{\bf W}\bfh,\alpha_0,\beta)$ \cite{barndorff1979edgeworth}.

As we have noted, the
surrogate density has a close relationship to $p_{0,x}(\bfz)$,
with $\bfh_z$ being the saddlepoint for the SPA of $p_{0,x}(\bfz)$.
%The SPA approximation is used when no closed-form expression exists
%for a distribution, but a closed-form expression for the moment generating
%function (MGF) is available \cite{BagNutKay2000}.
%The SP approximation is a first order inversion of the MGF, 
%and rapidly becomes very accurate as $N$ becomes large.
%Applying the SP approximation requires solving the SP equation to find the SP.
This follows from the fact that (\ref{tm1}) is 
equivalent to the {\it saddle point equation}
for the distribution of the linear sum of independent random variables
(i.e. see , equation 25 in \cite{BagNutKay2000}, on page 2245).
%To make the equivalence clear, the correspondence of variables between this paper and \cite{BagNutKay2000}
%are given by $\bfh\leftrightarrow\blambda$, ${\bf W}\leftrightarrow[a_{m,n}]$, $\balpha\leftrightarrow[b_n(\blambda)]$.
%It is also true that the theoretical activation function $\lambda(\balpha)$ 
%is identical to the derivative of the cumulative generating function (CGF)
%for the element distribution $p_e(x;\alpha,\alpha_0,\beta)$,
%denoted by $c^\prime( b_n(\blambda))$ in \cite{BagNutKay2000}.
%This follows from the property that the first derivative of the CGF 
%is identical to the expected value.

\subsection{Asymptotic Network and Deterministic PBN}
\begin{figure}[h!]
  \begin{center}
    \includegraphics[width=3.5in]{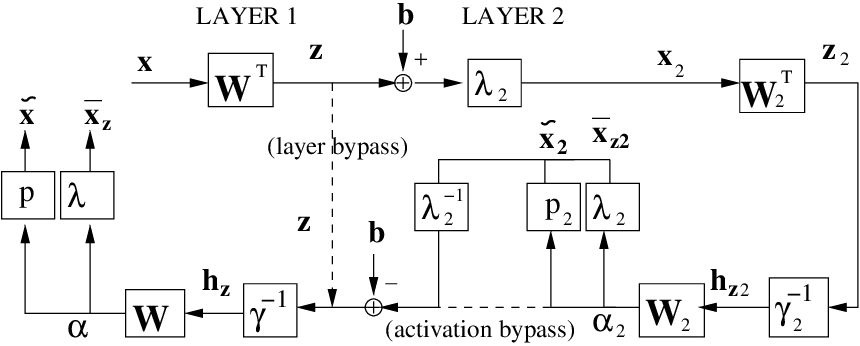}
  \caption{Block diagram of 2-layer PBN in asymptotic form.
Several versions of PBN are shown in one figure, as
implemented by dotted lines (see text).}
  \label{asy}
  \end{center}
\end{figure}

The surrogate density can be seen as a layer
in a generative network taking
a familiar form.  We first simplify notation by defining the function
$\gamma(\bfh)  =  {\bf W}^\prime \lambda \left( {\bf W} \bfh \right).$
By definition,  $\gamma(\bfh_z) = \bfz$.  We also define the inverse : $\bfh_z = \gamma^{-1}(\bfz).$
The concept of $\gamma^{-1}(\bfz)$ is illustrated in Figure \ref{asy}
in which a two-layer network is shown.  On the top of the figure is the forward
path, and on the bottom is the reconstruction path that uses the asymptotic form of the
network. We concentrate on just the first layer by using the shortcut path ``(layer bypass)".
Feature $\bfz$ is converted to $\bfh_z$ through $\gamma^{-1}(\bfz)$, then multiplied by ${\bf W}$ to raise the dimension
back to $N$, and finally passed through activation function $\lambda(\;)$ to produce $\bar{\bfx}_z$.
Optionally, it can be passed to the generating distributions $p_s(\bfx; \balpha,\alpha_0,\beta)$ for stochastic generation,
indicated by the block ``p" in the figure.
According to the definition of $\gamma^{-1}(\;)$, it is clear that  ${\bf W}^\prime \bar{\bfx}_z = \bfz,$
or in other words, the feature $\bfz$ is recovered exactly when  $\bar{\bfx}_z$ is processed
by the forward path (i.e., a right-inverse).  The generative (reconstruction) path
looks like the generative path of an autoencoder with tied weights, except the
activation function is replaced by $\gamma^{-1}(\bfz)$. 

This becomes more clear as we extend the network to two layers
by not using the shortcut path ``(layer bypass)".  The forward path (top of figure)
now passes through a bias and activation function before linear transformation
in the second layer.   The forward path is a standard
multi-layer perceptron, or ``feed-forward
neural network" (FFNN).  
The return path, although shown as a separate network in the figure is
tantamount to backing up through the FFNN.
With a second layer, something quite interesting happens
in the reconstruction path.
Similar to the first layer, as we begin the reconstruction process by applying the $\gamma^{-1}(\bfz)$
of the second layer to obtain $\bfh_{z2}$, then apply linear transformation
by matrix ${\bf W}_2$, we have the choice of
stochastic reconstruction through block ``$p_2$" or deterministic reconstruction using $\lambda_2(\;)$.
The result is a reconstruction of $\bfx_2$, the output of the first layer in the forward path.
It is a right-inverse reconstruction.
To continue to the left, the activation function and bias at the output of the first layer need to be inverted
in order to obtain a reconstruction of $\bfz$.  But, for deterministic
reconstruction, this inversion of the activation function cancels the application of $\lambda_2(\;)$
as long as the activation function is the same.
Therefore, using the reconstruction shortcut ``(activation bypass)"
eliminates the need for inversion of the activation function.
For stochastic generation, the activation function
is not used, so the cancellation with its inverse
does not occur, and this shortcut is not possible.

When entirely using deterministic reconstruction and ``(activation bypass)",
 the multi-layer reconstruction path looks like
a perceptron network, but the non-linearities are replaced by the functions
$\gamma^{-1}(\bfz)$, which do not operate element-wise.
This type of network is called deterministic PBN (D-PBN) \cite{BagEusipcoPBN,BagEusipco23TCA}.
We have called this network a ``network based on first principles" \cite{BagIcasspPBN}
because it lends insight into how an auto-encoder would be designed
if it were not influenced by classical neural network topology.
Furthermore, the network is a right-inverse network, so is a two-way network.

\subsection{Solving the Saddle Point Equation and Sampling Failure}
%[Felix: add a paragraph on how the SPA is applied to solve (\ref{tm1}) here.]
%
Solving (\ref{tm1}) presents a computational challenge
due to the need to repeatedly invert matrices of the size $M\times M$, where
$M$ is the output dimension of a layer \cite{BagNutKay2000}.  The problem can be 
avoided or mitigated by various methods \cite{BagPBNHidim}, 
some of which will be employed in the experiments.

Another potential problem with sampling a D-PBN is {\it sampling failure.} 
The solution to (\ref{tm1}) is not guaranteed to exist
unless $\bfz={\bf W}^\prime \bfx$ for some $\bfx$ in 
$\mathbb{X}$ \cite{BagIcasspPBN}. When backing up through
a FFNN, this condition is only strictly met for the last layer
(first layer when working backwards), meaning that some input samples cannot be auto-encoded.
Luckily, the {\it sampling efficiency}, which is the fraction of samples that succeed, 
can be driven close to 1 with proper initialization. Then
further training typically drives the sampling efficiency
to one, resulting in no sampling failures in the training set,
or in the validation set \cite{BagIcasspPBN,BagPBNEUSIPCO2019,PBNTk}.

Note that sampling a stochastic PBN is also subject to 
the problem of failed samples and sampling efficiency, 
but training a stochastic PBN is not because the likelihood function can be calculated for any sample.
Therefore, a D-PBN can be initialized by first training the network
as a PBN, then later as a D-PBN.  A far more efficient way to initialize a
D-PBN is to use the up-down algorithm, which we mentioned in Section \ref{dg1sec}. 
This can initialize a D-PBN with sampling efficiency equal to or near 1.
Then, D-PBN training can proceed if the contributions of failed
samples are weighted by 0. The sampling efficiency then rises. 
To understand why, note that sampling occurs on a manifold (\ref{mandef}), the drawn
%[Felix: add figure here ...]
sample of $\bfx$ is the conditional mean of $\bfx$ given $\bfz$, which is also
the weighted centroid of the manifold.  
This can be seen as the ``safest" solution, 
as far from the boundaries as possible, making it less likely that a sample 
will fail.  Sampling can also fail for some distorted or unusual samples,
which are poorly reconstructed anyway by other types of auto-encoders.

\subsection{Generative Classifier Topologies using PBN}
\label{topsec}
We now discuss ways to construct generative
classifiers, and how these topologies can be implemented
using PBN, and which approach is preferred. 
Consider data vector $\bfx$ and 
data class labels $i\in\{1,2 \ldots K\},$
having prior class probabilities $P(i)$.
From the discrete labels, we can construct 
a label signal $\bfy$ that can take one of $K$
values $\bfy \in \{\bfy_1,\bfy_2 \ldots \bfy_K\},$
suitable for processing by a network,
using one-hot encoding or other methods.
We consider the following three main approaches to constructing
classifiers:
\benum
   \item {\bf Discriminative approach.}
In the discriminative approach, the task is to directly estimate
the conditional label probabilities  $P(i|\bfx)$
or the posterior label signal densities $p(\bfy|\bfx)$.
This is the standard approach used in classifiers
because it is direct and does not require estimating
probabilty distributions of the input data.
   \item {\bf Conditional distributions (generative).}
The simplest and most straight-forward generative classifier
topology is the so-called Bayesian classifier
constructed from the class-conditional data distributions
$p(\bfx|i)$, which we also write as $p_i(\bfx)$.
The label probabilities are calculated 
as a second step using Bayes rule:
$P(i|\bfx) = \frac{p(\bfx|i)\;P(i)}{p(\bfx)},$
however the normalizing factor $p(\bfx)$ is unnecessary
if classification is implemented using
$$\arg\max_i\{p(\bfx|i)\;P(i)\}.$$
Decision boundaries are formed by the comparison
of likelihood function values.
To implement this with a PBN, we train
a separate PBN on the training data for each class
to obtain $p(\bfx|i)$.  The advantage of training a separate model on 
each class is the large class-selectivity that it can
provide.  By training on each class separately, network weights
become more fine-tuned to a given class. In convolutional
layers, convolutional kernels take on patterns 
that act as basis functions for representing data of the
given class.  This approach , however,
has one very important disadvantage: training a separate
model on each class allows differences in initialization and training to
greatly influence the resulting models and cause imbalances,
which shift the decision boundaries, causing classification errors.
There are ways to mitigate this. For example
in speaker recognition, GMM-UBM \cite{bhattacharjee2012gmm}
uses a single Gaussian mixture model (GMM) probability density
estimate as a base model (the universal background model - UBM)
which is then modified for each data class using just 1 step
of the E-M algorithm.  This eliminates the effects 
of randomly initializing each model.
The general idea can be used when training PBNs.
Class-dependent PBNs can be trained starting with a common
model trained on all classes, but the benefits of this
are limited and it can reduce class selectivity.
   \item {\bf Joint distribution (generative).}
The joint-distribution approach trains a single model on the
joint distribution of the class and label, $p(\bfx,\bfy)$,
then classifies using $$\arg\max_i\{p(\bfx,\bfy_i)\}.$$
Although it avoids the model mismatch problem
of the conditional distributions approach,
it has one important flaw. Generative models 
are only approximations to the
true distribution. In any approximation, compromises are made
between conflicting goals: model simplicity and model accuracy.
Since we intend to use joint distribution $p(\bfx,\bfy)$ as a classifier, 
and there usually is insufficient data to accurately estimate 
the true distribution, we must compromise by using a simplified
model.  We hope that the compromise favors high class selectivity,
but we cannot guarantee this, nor can it
be forced.  In short, the joint-distribution approach
avoids the disadvantages
of the conditional distributions approach, but does not have its
advantages, nor does it have the advantages of
the discriminative approach.  

% This is especially true in a PBN
%network consisting of multiple layers.
% The higher the dimension a layer's input data
%has, the greater the effect of changing the layer weights
%has on the cost function.  Weights in convolutional
%kernels near the input of the network
%will strive for general
%
%Therefore, the labels will have less
%effect if they interact with weights at the network output. 
%layers, 
%
%
%
Furthermore, a problem also arises because $\bfx$ and $\bfy$ have different forms,
so the complete input data $\{\bfx,\bfy\}$ is non-homogeneous.
This is especially true when $\bfx$ is 
multi-dimensional (i.e.  an image or spectrogram).  
It is not clear how to insert $\bfy$ into the input
image, or combine it with the image so that it can be 
meaningfully processed by a convolutional network. 
This problem can be avoided by injecting the label signal
$\bfy$ into intermediate convolutional layers
after some down-sampling, or into dense (fully-connected) 
layers near the end of the network.
Examples of injecting labels into the data
can be found in design of class-dependent GANs \cite{MiyatoGAN2018}
or in a deep belief network (DBN) \cite{HintonDeep06}.

The joint distribution approach has a further disadvantage
when using PBNs.  
In a PBN, a kind of vanishing-gradients problem arises 
with respect to the labels. The higher the dimension a layer's input data
has, the greater the effect of changing the layer weights
has on the cost function.  Therefore, the labels will have less
effect if they interact with weights at the network output. 
%
%But, as a natural result of back-propagation in a deep
%network, injecting label signals at the end of the network
%reduces the influence of the labels on the
%layer weights at the input layers.  As a result, the 
%initial convolutional weights
%may not be sufficiently coaxed to extract
%discriminative information.
%
%of the hidden variable,
%the larger the influence on the likelihood function.
%
As an aside, injecting label information in a PBN at multiple layers
is not justified theoretically because the
likelihood function for such a network may be intractable.
In short, the joint distribution approach is problematic for PBN and 
likely to suffer from poor class-selectivity
in the likelihood function.

%   \item {\bf Mixture Distribution.}
%It is possible to mitigate the problems of 
%the conditional distributions method using
%ideas of mixture distribution, similarly
%to the approach explained in Section \ref{pdfpmultif}. 
%
%To add mathematical detail, consider $L$ feature extraction chains, resulting in features $\bfz_1, \ldots \bfz_L$, 
%with feature probability density functions $g_1(\bfz_1), \ldots g_L(\bfz_2)$. 
%Applying (\ref{ppt0}) separately to each model, we arrive
%at the projected PDFs $G_1(\bfx), \ldots G_L(\bfx)$. 
%The mixture distribution can be created as follows
%$$G(\bfx) =  \sum_{i=1}^L  \;\alpha_i \;G_i(\bfx).$$
%Such a mixture model is then trained separately on each class, with the weights $\{\alpha_i\}$ also condidered
%parameters to estimate.
%
\eenum
To summarize, we prefer the conditional distributions
approach for a PBN classifier because it has
 the potential for high class selectivity as a result
of training networks individually on each class. 
But, we must seek ways to avoid the problem of model imbalance
and the resulting shift of decision boundaries.
This problem is solved by discriminative alignment.

\section{Discriminative Alignment of PBN (PBN-DA)}
\label{sndbx}
For reasons mentioned in Section \ref{topsec},
it is generally assumed that the discriminative approach to classification
is superior to the generative approach.
In short, estimating the class-dependent data
 distributions at high dimension is much harder than
predicting the class label \cite{Vapnik99,Goodfellow2016}.  
But, because a PBN is a two-directional network
one should be able to implement both paradigms in one network.
But how? There appears to be a contradiction: 
discriminative classifiers are trained simultaneously on all 
data classes, but the conditional distributions approach,
which we have said we want to use (see Section \ref{topsec}),
requires training separate networks on each class.
%\textcolor{red}{
%There is one problem with the claim that a PBN is simultaneously discriminative and generative:
%a discriminative classifier is trained on all data classes, 
%while purely generative classifier (that is to say a Bayesian classifier)
%must be trained separately on each class. 
%}
%\textcolor{green}{
%}
A solution to this dilemma was proposed by the method of
discriminative alignment \cite{BagSPL2021,BagPBNHidim,BagEusipcoESC8} in which each
class-dependent network is trained simultaneously as a generative
model for the given class, but also as a discriminative model against ``all other classes".
This approach tends to ``align" the network weights, giving the generative
model high selectivity against the other classes, 
thus getting the best of both the generative and discriminative approaches.
By causing high selectivity against other classes, it
mitigates the problem of model mismatch because
with high selectivity (i.e. high likelihood function
slope when moving in the direction of other classes), so the decision boundaries
will not shift much due to model mismatch.

\begin{figure}[h!]
\begin{center}
	\includegraphics[width=3.4in, height=0.9in]{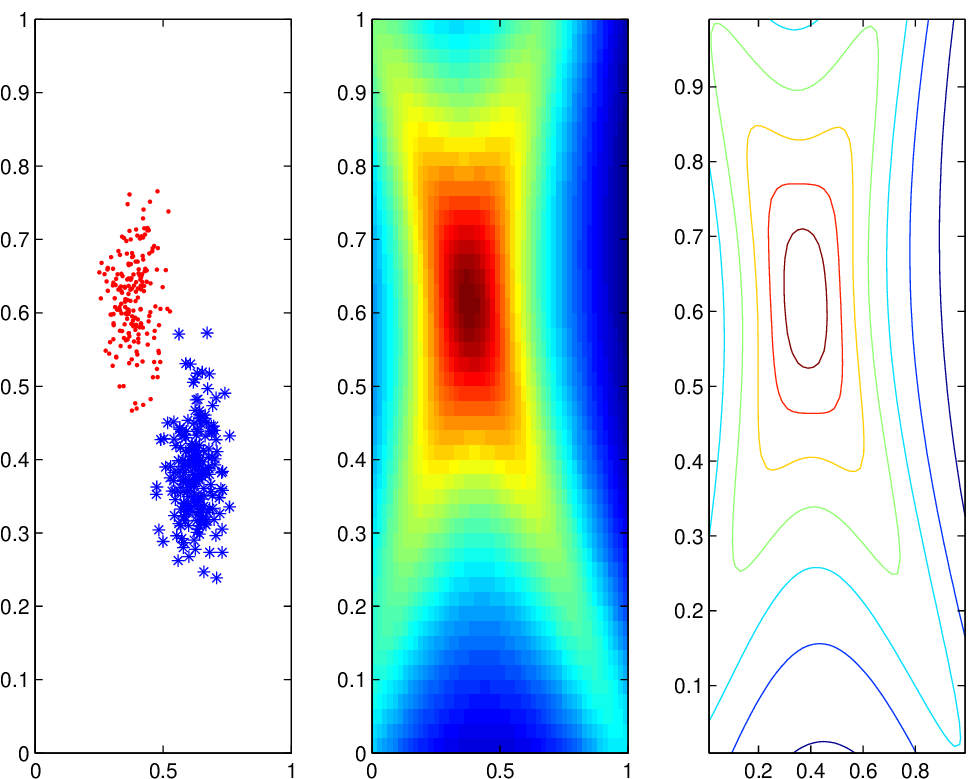}
	\includegraphics[width=3.4in, height=0.9in]{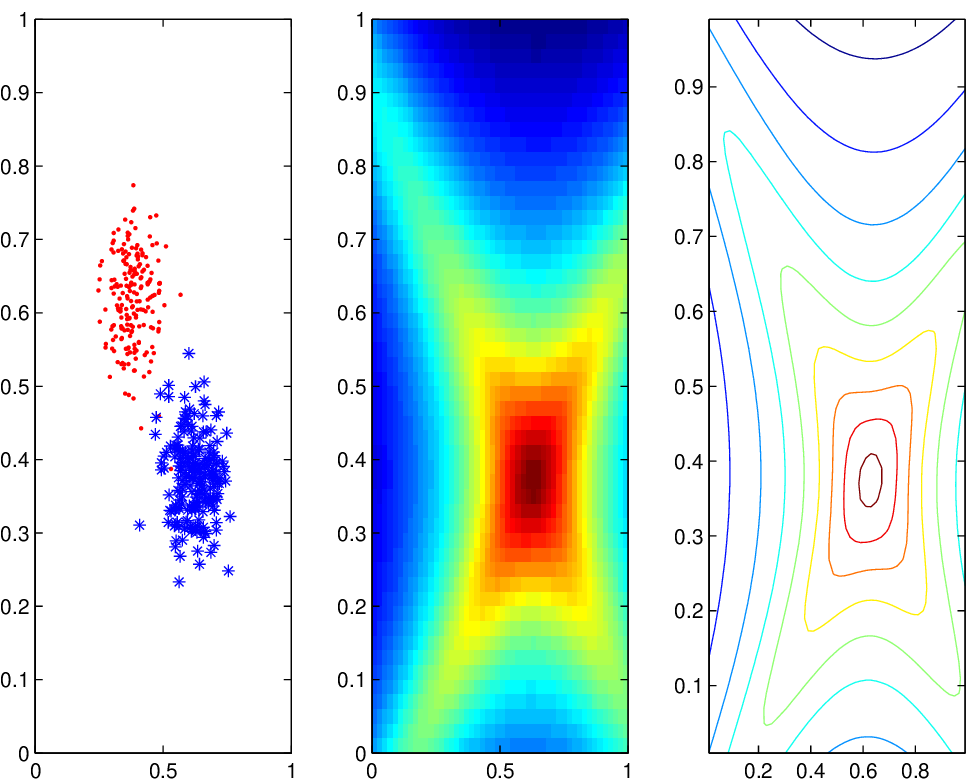}
	\includegraphics[width=3.4in, height=0.9in]{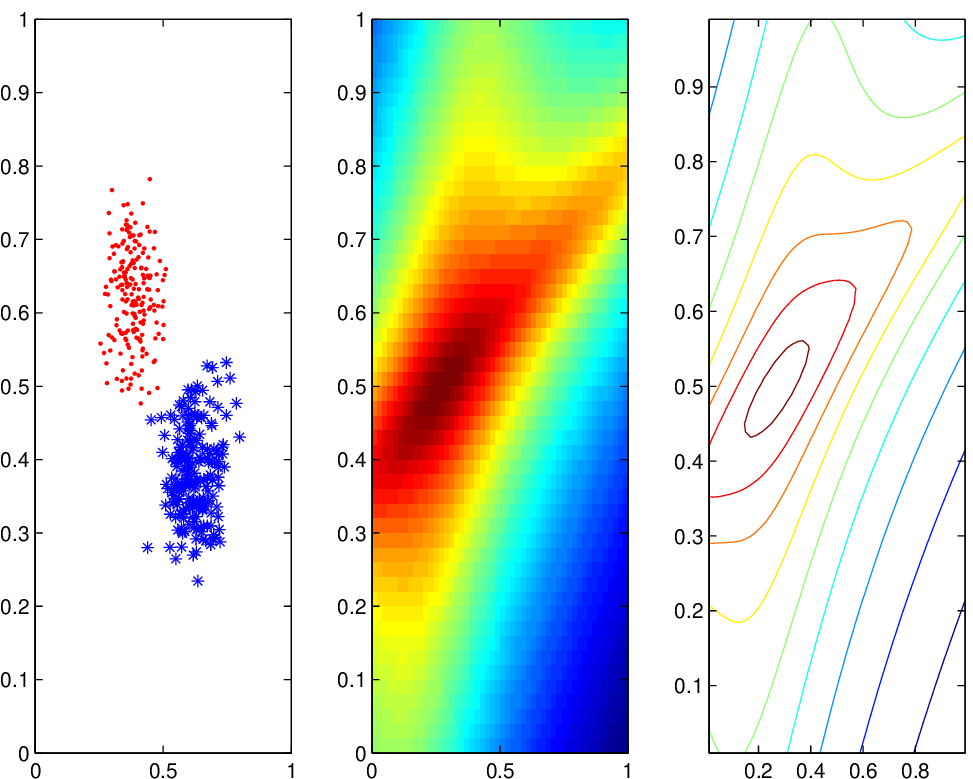}
	\includegraphics[width=3.4in, height=0.9in]{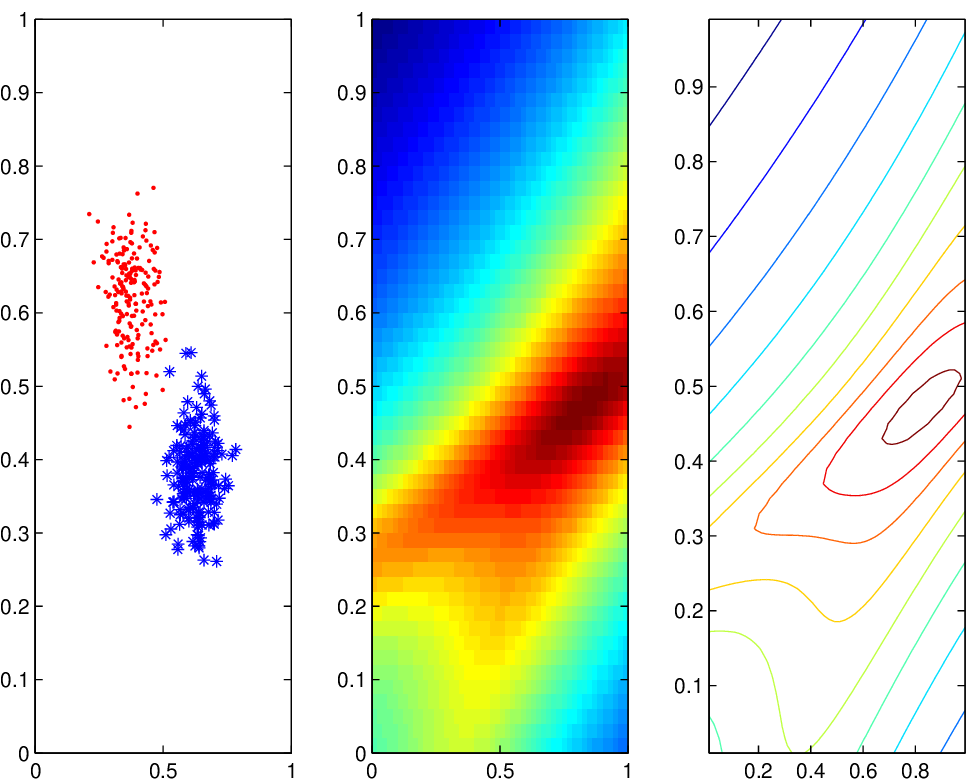}
	\caption{From top to bottom: PBN trained on class 1 (red), 
PBN trained on class 2 (blue), PBN trained on class 1 with
discriminative alignment, PBN trained on class 2 with
discriminative alignment. Left column: input data, center: likelihood
surface, right: contour lines of likelihood surface.}
	\label{pdf_c}
\end{center}
\end{figure}

To see this visually, we trained a simple PBN network
on two-dimensional data with two data classes.
In Figure \ref{pdf_c}, on the top row, we see
data from the two classes (left), an intensity plot of the
PBN likelihood function after training on the first data class
(center), and the corresponding likelihood contours (right).
As we would expect, the likelihood has a peak at the location 
of the data. On the second row, the analogous
plots are seen for a network trained on the second data class. 
However, as can be seen by the contour plots, there is not much selectivity
against the other data class because the slope is highest in 
the orthogonal direction.  In short, the class-selectivity 
appears to be poor.

We then re-trained the two
PBNs with combined cost function. A classifier (cross-entropy)
cost component was added to the PBN log-likelihood cost,
giving the network the additional task of discriminating
the two classes. In rows 3 and 4 of the figure,
we see the results. Interestingly, now the contour plots show
high selectivity against the other data class
(i.e. high slope in the direction that separates the two data classes).
The contours have been ``discriminatively aligned".
Now, when classifying class 1 vs. class 2 using a straight Bayesian
likelihood classifier, the classification results will
resemble the properties of the discriminative classifier.
In short, the best qualities of both network types
are realized.
The method has shown very promising results when compared
to state of the art discriminative classifiers \cite{BagSPL2021,BagPBNHidim,BagEusipcoESC8}.

\subsection{Self-Combination of PBN-DA}
\label{selfcomb_sec}
As we have claimed, the PBN is a two-directional network,
incorporating two distinct network topologies in a single embodiment.
In an $K$-class classifier scenario, the PBN-DA consists
of $K$ distinct networks, each trained on a single class.
The forward networks are trained as a classifier to discriminate one class
from all others, and the generative PBN likelihood function is
maximized over the training data of the same class.
It follows, that one could use two methods to 
construct a classifier from a trained PBN-DA (a) a straight
generative ``conditional distributions" approach classifier 
based on comparing the projected log-likelihood functions
of the $K$ PBNs, and (b) a forward classifier based on
finding the maximum output statistic across all
class assumptions and all networks.
We could therefore construct two distinct classifiers from
a single set of trained PBNs. 

Alternatively,  the discriminative
influence can be integrated into and combined with 
the generative classifier using a special
output distribution.  Notice that the last term in the
projected likelihood function for each PBN is the output feature
distribution, for example $g(\bfz)$ in (\ref{ppt1}).  
Let $g_m(\bfy)$ be the output feature distribution for the network trained
on class $m$.  Since $g_m(\bfy)$ is given, it can be 
constructed as an indicator function for class $m$, 
so that it has a higher likelihood if $\bfy$ equals
the label signal for class $m$ (e.g. one-hot encoding), 
given by $\bfy_m$.
For example, 
\beq
g_m(\bfy) = \delta[ \bfy - \bfy_m].
\label{gmg}
\eeq
In the experiments, we approximated (\ref{gmg})
with a TED ouput activation function
(See Table \ref{tab1a}) which, similar to 
sigmoid, has an output in the range $[0, 1]$.
We assume the TED distribution
$$g_m(\bfy)= \prod_{i=1}^K \; \left(\frac{\alpha_{i,m}}{e^{\alpha_{i,m}} - 1}\right)  \; e^{\alpha_{i,m} y_i},$$
where $\bfy=\{y_1 \ldots y_K\}$, and $\alpha_{i,m} = \{ -C, \; i\neq m, \;\; +C, i=m\}.$ 
As can easily be verified, $g_m(\bfy)$ will have high value when $\bfy$ 
is the one-hot encoding for class $m$,
and a low value otherwise.
The constant $C$ can be used to control the relative importance of the output statistic (i.e. the class labels) in the projected LF.
This has an effect similar to linear combining (ensembling) two classifier statistics (in one network!), and it is possible to experimentally determine the optimal value of $C$.

\subsection{PBN-DA-HMM}
\label{pbndahmms}
The hidden Markov model (HMM) \cite{RabinerHMM} was for a long time the state of the art in
acoustic event classification and modeling.  The HMM owes its success to two things:
\benum
\item The Markov assumption, which allows modeling events with unknown start time, and duration.
The HMM is very forgiving against time-distorted or intermittent time sequences, a common
problem in real-world speech and acoustic events.
\item The efficient forward algorithm for implementation of training and evaluation.
\eenum
With the advent of convolutional neural networks (CNNs), the detection and modeling of events with unknown
temporal location and size was greatly improved.  

Despite the success of CNNs, it remains a possibility that the HMM is still useful as a component of a PBN.  
The LF of a PBN is formed recursively from the contributions of each layer according to
the approach in equation (\ref{ppt1}).
Suppose a PBN is broken into two parts. Let the hidden variables coming out of the first half
be denoted by $\bfh$. Then, using PDF projection, the second half of the network
implements a PDF estimate $\bfh$, denoted by $G(\bfh)$, e.g. equation (\ref{ppt1}).
%
%And, when a PBN is broken in the middle at a given layer, the hidden variables $\bfh$,  
%can be considered the output of the shortened network, and the
%PBN composed of the remaining layers has a LF that implements
%a PDF estimate for $\bfh$, denoted by $p(\bfh)$.
However, it can be that the dimension of $\bfh$ is small enough that one
can use well-known PDF estimation methods, such as HMM,  to replace, 
and possibly improve upon $G(\bfh)$. 
%Typically in a convolutional neural network (CNN),
%the feature map at the last convolutional layer is flattened, and passed
%to a series of dense layers. This flattening approach, while practical, may not
%lead to the best representation of the feature distribution because
%it ignores the special meaning of the time dimension in acoustic events.
The Markovian assumption exploited by the HMM, while over-simplified, provides an excellent
compromise between tractability and generative modeling accuracy.
There are two advantages to this, (a) the use of
discriminative alignment in the pre-training of the
first part of the network lends discriminative information
to the features (class-selectivity), and (b) the use of HMM to replace the   
second part of the network has the stated advantages of the Markov model.

To summarize the method of PBN-DA-HMM: we train a full PBN 
as PBN-DA on class $m$, then when finished, shorten the network,
tapping off the intermediate features $\bfh_m$,
and estimate the distribution $p_m(\bfh_m)$ using HMM.
This is repeated for all classes $m$, and finally a 
class-specific classifier
is constructed using the method in Section \ref{pdfpmultif}.
%approach at the end of Section \ref{pdfprojsec} by multiplying 
%$p_m(\bfh_m)$ by the corresponding J-function (see beginning of Section \ref{pdfprojsec}).

\section{Experiment 1: ESC50 Data Set}

\subsection{Data Selection, Feature Selection, and Feature Extraction}
The environmental sound classification data set (ESC50) \cite{ESC50} consists of $50$ data classes, with
 $40$ ten-second air-acoustic recordings in each data class. 
The classes are diverse, and it is difficult to represent them well by 
one feature extraction approach alone.   As outlined in Section \ref{pdfprojsec},
PDF projection allows the use of multiple feature extraction approaches in 
a single common generative model.  

To this end, we attempted to match each data class with a feature
set. This was done by segmenting the time-series into overlapped
Hanning-weighted segments, then extracting 
log-MEL-band features. Time-series were then re-synthesized
by reconstructing the power spectrum from the log-MEL-band
features using maximum-entropy feature inversion \cite{BagUMS,BagIcasspPBN},
as implemented by $\bar{\bfx}_z$ in Section \ref{asysec},
then reconstructing the complex-valued FFT
using random-phase, reconstructing the segment time-series by
inverse-FFT, Hanning-weighting and
finally re-synthesizing time-series by using overlap-add.
The time-series were
played back and compared acoustically with the original.
For almost all classes, a segment size and number
of MEL bands could be found to result in very good reconstruction,
as determined subjectively by ear.
A large number of classes (23 classes)
were well adapted to a segment size of
768, 2/3 overlap,  and 48 log-MEL-spaced bands.
The 23 classes (classes are numbered
0-49) were 0,1,6,9,10,11,13,16-19,21,23-25,27-30,36,45,47,49.
The features for each 10-second event consisted of
624 time samples, resulting in a 624$\times$48 (time $\times$ freq)
matrix.  

As explained in Section \ref{pdfpmultif}, PBN classifiers
can be constructed using multiple feature sets.
But, this would require significantly more
time and processing resources and would
detract from evaluating the benefits
of discriminative alignment together with HMM.
Therefore, it was decided to put off any multiple-feature-set
experiments using all 50 classes to future work
and conduct a limited feasibility experiment
on the 23-class subset using one feature type.  

In a previous publication \cite{BagEusipcoESC8}, we reported the results
of experiments for an 8-class subset of these same 23 classes.
We now provide results for all 23 classes.
There were then 23$\times$40 = 920 total events.

%\subsection{Data Partitioning}
To partition the data, we used 4:1 random data holdout,
%\textcolor{red}{why? we should justify why we do not simply use the predefined dataset splits, which ensures that individual samples are not from the same recordings and would ensure that results are reproducible: \emph{The dataset has been prearranged into 5 folds for comparable cross-validation, making sure that fragments from the same original source file are contained in a single fold.}} 
selecting 10 testing samples of the 40 samples of each class at random,
and trained on the remaining 30.  
We did this four times, independently.
There were 230 testing samples in each fold.
The partitions were designated by letters A-D.
For reproducibility, we provide these features and data folds
online \cite{CSFTk}.

\subsection{Network Architectures}
We used two network architectures in the experiments, 
(a) the ``PBN networks", which were trained separately 
on each of the 23 classes, and (b) a state of the art ``CNN classifier"
trained jointly as a classifier on all classes. 
\\

\noindent
{\bf PBN networks}. We used a network similar to that used in \cite{BagEusipcoESC8}.  The eight-layer PBN network had three convolutional and 5 dense layers, ending with a classifier layer
of 23 neurons. The input data is 624$\times$48 (time $\times$ freq).
Kernels in the three convolutional layers were 8 12$\times$16 kernels, 
30 10$\times$5 kernels, and 120 21$\times$3 kernels
respectively. Downsampling was 3$\times$4, 3$\times$2 , and 3$\times$1. 
Convolutional border modes were ``valid".  The dense layers had 512, 
256, 128, 128, and 23 neurons. The last layer is the cross-entropy 
classifier (output) layer.  The output of the third convolutional 
layer has dimension 16$\times$120, which is tapped off for HMM processing.
For the HMM, the data is seen as having 16 time steps and a feature 
dimension of 120.  The twenty-three class-dependent PBN networks used linear 
activation at the output of the first 3 layers,
in order to form a Gaussian group \cite{BagPBNHidim}, and thereby
greatly reducing the required computation. The remaining 
layers used the truncated Gaussian (TG) activation \cite{Bag2021ITG}, 
similar in behavior to softplus, not unlike
leaky Relu \cite{maas2013rectifier}, but continuous (see Table \ref{tab1a}).
\\

%\noindent
%{\bf DNN classifier}.  For the DNN classifier, we used
%the same network as for the PBN networks, but used TG activation function in all layers, 
%and max-pooling instead of down-sampling in the convolutional layers.
%We also used dropout regularization in layers 4 and 5.

\noindent
{\bf CNN classifier}.
\begin{table}[!t]
\centering
\caption{Modified ResNet architecture of the CNN.}
\begin{adjustbox}{max width=\columnwidth}
\begin{tabular}{lll}
\toprule
layer name & structure & output size\\
\midrule
input & BN (temporal axis) & $624\times48$\\
2D convolution & $7\times7$, stride$=(2,1)$ & $312\times48\times16$\\
residual block & $\begin{pmatrix}3\times3\\3\times3\end{pmatrix}\times 2$, stride$=1$ & $155\times46\times16$\\
residual block & $\begin{pmatrix}3\times3\\3\times3\end{pmatrix}\times 2$, stride$=1$ & $78\times23\times32$\\
residual block & $\begin{pmatrix}3\times3\\3\times3\end{pmatrix}\times 2$, stride$=1$ & $39\times12\times64$\\
residual block & $\begin{pmatrix}3\times3\\3\times3\end{pmatrix}\times 2$, stride$=1$ & $20\times6\times128$\\
max pooling & $20\times1$, stride$=1$ & $1\times6\times128$\\
flatten & BN & $786$\\
dense (embedding) & linear & $128$\\
sub-cluster AdaCos & $4$ sub-clusters per class & $23$\\
\bottomrule
\end{tabular}
\end{adjustbox}
\label{tab:cnn}
\end{table}
The CNN model consists of a modified ResNet architecture \cite{he2016residual} with about $800$ k trainable parameters and is shown in Tab. \ref{tab:cnn}.
It consists of an initial temporal mean normalization operation, followed by a convolutional layer, 4 residual blocks, a temporal max-pooling operation and a dense layer with a linear activation function.
The residual blocks each consist of two convolutional layers with kernels of size $3\times3$, batch normalization, a max pooling operation of size $2\times 2$ and use the ReLU activation function.
Futhermore, the whole network does not contain any bias terms.
The same architecture has been used successfully for the 
8-class subset of ESC-50 \cite{BagEusipcoESC8}, few-shot open-set acoustic event classification \cite{wilkinghoff2023using} and anomalous sound detection \cite{wilkinghoff2023design}.
The model learns to embed audio data onto a hypersphere of fixed dimension and is trained by minimizing the angular margin loss sub-cluster AdaCos \cite{wilkinghoff2021sub}.
This loss is an angular margin loss with an adaptive scale parameter similar to AdaCos \cite{zhang2019adacos} but uses multiple instead of a single centers for each class, called sub-clusters.
In this work, we used an embedding dimension of $128$ and $4$ randomly initialized sub-clusters for each class that are not adapted during training.
When training the model, dropout with a probability of $50\%$ is applied to the hidden representations before the last dense layer \cite{hinton2012improving}. Additionally, two data augmentation techniques were used to improve the performance of the CNN.
As a first technique, mixup \cite{zhang2017mixup} was applied to the input samples and their corresponding classes using a mixing coefficient sampled from a uniform distribution.
Secondly, we applied random shifts up to $10\%$ of the temporal dimension of the input representations
(which was 640 time steps, so shifts were +/- 64 samples).
The CNN is trained for $500$ epochs with a batch size of $8$ using adam \cite{kingma2015adam} and is implemented in Tensorflow \cite{abadi2016tensorflow}.
To have a fair comparison, no external data was used to train the CNN.
\par
After training, each test sample is embedded into the embedding space by applying the mapping learned by the CNN and projected to the unit sphere by normalizing with respect to the Euclidean norm.
Then, similarity scores for each class are computed by taking the cosine similarity to the class-wise mean embeddings of all normalized embeddings extracted from the training samples.
The class of the mean embedding with the shortest distance is used as the classification result.

%\subsection{CNN-AUG}
%Additionally, we used two data augmentation techniques to improve the performance of the CNN.
%\textcolor{red}{ For comparison of methiods, when one used DA and the other not, it is unfair.  }
%As a first technique, we applied mixup \cite{zhang2017mixup} to the input samples and their corresponding classes using a mixing coefficient sampled from a uniform distribution.
%Secondly, we applied random cyclic temporal shifts to the spectrograms.
%The resulting model is denoted with CNN-AUG.
%Applying these techniques slightly improves the performance of the CNN model (see. %Tab. \ref{tab1r}), but not significantly.

\subsection{Network Training and Initialization}
\label{pbn_init_sec}
A separate PBN was trained on each of the 23 classes using
discriminative alignment. About 1500 epochs were required.
In addition to looking for convergence
of the PBN LF, we also sought to have zero training errors
in the discriminative task of ``class $m$" against ``all other
classes".   The cross-entropy discriminative cost function
was scaled by 1000 before subtracting from the LF.

These networks were then used as-is for PBN-DA,
then shortened and used for PBN-DA-HMM
as described in Section \ref{pbndahmms}.
%To summarize the method of PBN-DA-HMM: we train a full PBN 
%as PBN-DA on class $m$, then when finished, shorten the network,
%tapping off the intermediate features $\bfh_m$,
%and estimate the distribution $p_m(\bfh_m)$ using HMM, then finally
%create a generative classifier using the class-dependent feature
%approach at the end of Section \ref{pdfprojsec} by multiplying 
%$p_m(\bfh_m)$ by the corresponding J-function (see beginning of Section \ref{pdfprojsec}).
%(a) We first train a separate PBN on each class $m$ using a combined
%cost function consisting of negative log-likelihood PBN cost, and cross-entropy 
%classification cost for the given class against ``all others", scaled
%by a factor of 1000, (b) 
The networks were shortened by tapping off
the output of the third layer, with data shape 16$\times$120.
We trained an HMM to estimate the probability distribution $p_m(\bfh_m)$ of this output map.
We then added $\log p_m(\bfh_m)$ to the combined log-J-function for the first 3 layers
to obtain the projected input data LF for PBN-DA-HMM.

%,
%(d) using PDF projecton, project this feature likelihood function back to the input data space by 
%adding combined log-J-function for the first 3 layers to $\log p_m(\bfh_m)$.
%As for PBN-DA, the classification is made by choosing the PBN network with highest projected log-likelihood.
%An example of this for a 2-layer network is equation (\ref{ppt1}), where $g(\bfz)$, 
%is replaced by the HMM distribution.
For data augmentation during training, we used random circularly-indexed
time shifts with a maximum of $\pm 40$ time segments.
Because the CNN and HMM were significantly dependent on the random initialization,
we always conducted three trials, and averaged the results.
Due to computational expense, it was not practical to repeat the training of the PBN networks
over multiple trials, but the four-fold partioning provided
adequate statistical diversity.

Note that the networks obtained by step (a) are used for PBN-DA in the experiments,
and from step (d) for PBN-DA-HMM.

\subsection{Computational Requirements for PBNs}
Using PBN Toolkit software \cite{PBNTk}, with 
a GPU (NVIDIA GeForce GTX 1050 Ti), and double pecision
(float64) with a batch size of 230, we are able to compute
one epoch in 59 seconds.  Double precision was needed
to avoid problems during inversion of matrices. 
It typically required about 1500 epochs to train each of the
23 models, equating to about one day per class.  
With two computers, the training was finished in less than 2 weeks.
We used stochastic gradient descent (or gradient ascent for LF),
and ADAM optimization algorithm.  
Note that when training one model, we use data from all classes, not just
the corresponding class. However, only data from the corresponding class 
is applied to the gradient of the LF cost function, whereas data from all classes is 
applied to the gradient of the discriminative cost function.

\subsection{Self-Combination of PBN-DA}
We experimentally determined the optimal value of parameter $C$,
as explained in Section \ref{selfcomb_sec}, by evaluating the number
of classification errors for PBN-DA, averaged over the four folds.
\begin{figure}[h!]
\begin{center}
	\includegraphics[width=2.0in, height=1.6in]{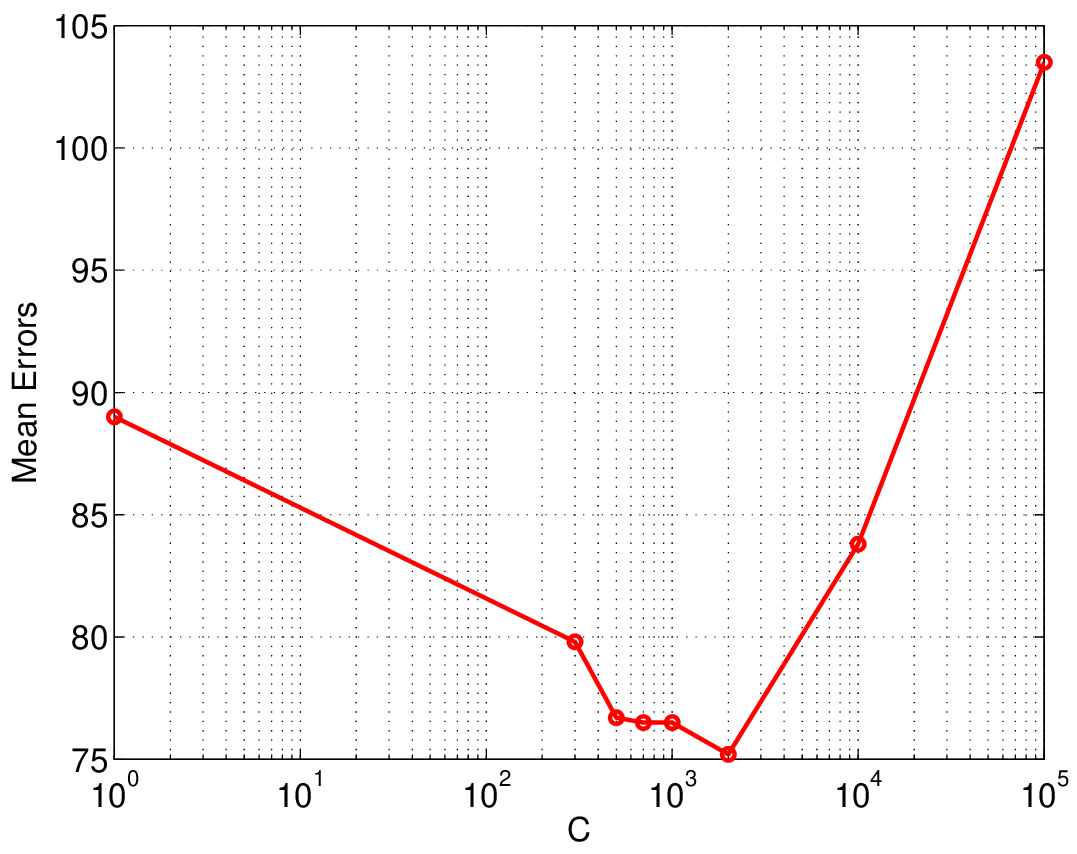}
	\caption{Mean errors for self-combination as a function of $C$.}
	\label{selfcomb}
\end{center}
\end{figure}
As can be seen in Figure \ref{selfcomb}, there is a distinct minimum at about
$C=2000$.  The value $C=2000$ was used in subsequent experiments for PBN-DA. 

%>> combine_kevin23
%Partition a, trial 1, CNN 29, PBNC-HMM 20, best comb: 18
%Partition a, trial 2, CNN 36, PBNC-HMM 31, best comb: 23
%Partition a, trial 3, CNN 36, PBNC-HMM 30, best comb: 19
%Partition b, trial 1, CNN 29, PBNC-HMM 21, best comb: 13
%Partition b, trial 2, CNN 27, PBNC-HMM 20, best comb: 9
%Partition b, trial 3, CNN 31, PBNC-HMM 18, best comb: 11
%Partition c, trial 1, CNN 28, PBNC-HMM 17, best comb: 14
%Partition c, trial 2, CNN 23, PBNC-HMM 20, best comb: 9
%Partition c, trial 3, CNN 30, PBNC-HMM 7, best comb: 10
%Partition d, trial 1, CNN 27, PBNC-HMM 59, best comb: 15
%Partition d, trial 2, CNN 23, PBNC-HMM 55, best comb: 18
%Partition d, trial 3, CNN 29, PBNC-HMM 54, best comb: 17
%>> cd ../ESC23full/
%>> combine_kevin23_full
%Partition a, trial 1, CNN 29, PBNC 77, best comb: 30
%Partition a, trial 2, CNN 36, PBNC 77, best comb: 27
%Partition a, trial 3, CNN 36, PBNC 77, best comb: 28
%Partition b, trial 1, CNN 29, PBNC 88, best comb: 27
%Partition b, trial 2, CNN 27, PBNC 88, best comb: 26
%Partition b, trial 3, CNN 31, PBNC 88, best comb: 26
%Partition c, trial 1, CNN 28, PBNC 84, best comb: 27
%Partition c, trial 2, CNN 23, PBNC 84, best comb: 24
%Partition c, trial 3, CNN 30, PBNC 84, best comb: 25
%Partition d, trial 1, CNN 27, PBNC 52, best comb: 28
%Partition d, trial 2, CNN 23, PBNC 52, best comb: 24
%Partition d, trial 3, CNN 29, PBNC 52, best comb: 27

\subsection{Individual Results}
\begin{table}
    \caption{Number of errors (out of a total if 230)
on each of the 4 data partitions for the various classifiers.  }
    \begin{center}
        \begin{tabular}{|l|l|l|l|l|l|l|}
            \hline
            & & \multicolumn{4}{|l|}{Partition} & \\
            \hline
            Algorithm & Trial & A & B & C & D & mean\\
            \hline
            CNN        & 1   & 29 & 29 & 28 &   27 & 28.2 \\
            CNN        & 2   & 36 & 27 & 23 &   23 & 27.2\\
            CNN        & 3   & 36 & 31 & 30 &   29 & 31.5 \\
            \hline
            CNN        & mean& 33.7 & 29 & 27 & 26.3 & {\bf 29}\\
       	    \hline
            PBN-DA & n/a   & 77 & 88 & 84 &   52 & {\bf 75.25 }\\
       	    \hline
            PBN-DA-HMM & 1   & 20 & 21 & 17 &  59 & 29.3 \\
            PBN-DA-HMM & 2   & 31 & 20 & 20 &  55 & 31.5 \\
            PBN-DA-HMM & 3   & 30 & 18 & 7 &   54 & 27.3 \\
            \hline
            PBN-DA-HMM & mean& 27 & 19.7 & 14.7 & 56.0 & {\bf 29.3} \\
            \hline
            CNN+PBN-DA & 1   & 30 & 27 & 27 &   28 & 28\\
            CNN+PBN-DA & 2   & 27 & 26 & 24 &   24 & 25.2 \\
            CNN+PBN-DA & 3   & 28 & 26 & 25 &   27 & 26.5\\
            \hline
            CNN+PBN-DA & mean& 28.3 & 26.3 & 25.3 & 26.3 & {\bf 26.6}\\
       	    \hline
            CNN+PBN-DA-HMM & 1   & 18 & 13 & 14 &   15 & 15\\
            CNN+PBN-DA-HMM & 2   & 23 & 9 & 9 &   18 & 14.7 \\
            CNN+PBN-DA-HMM & 3   & 19 & 11 & 10 &   17 & 14.2\\
            \hline
            CNN+PBN-DA-HMM & mean& 20 & 11.0 & 11 & 16.7 & {\bf 14.7}\\
            \hline
            %PBN-DA-HMM and CNN\tablefootnote{These numbers are obtained using an optimal combination factor with a value of ??? in this case.} &  &  &  &  &  \\
            %\hline
        \end{tabular}
    \end{center}
     \label{tab1r}
\end{table}

The number of errors out of a totlal of 230 events
in each of the 4 data partitions are shown in Table 
\ref{tab1r} for \ac{cnn}, PBN-DA, and PBN-DA-HMM classifiers. 
Three random trials are shown for all methods
except PBN-DA. Due to computational expense, the PBNs are trained just once.
For PBN-DA-HMM, three independent trials pertain just to the HMM.

First, it can be seen that the CNN classifier has significantly
better performance than PBN-DA.  In previous publications \cite{BagSPL2021,BagPBNHidim,BagEusipcoESC8},
PBN-DA competed more favorably with CNN, but it must be pointed out that in these
experments, CNN used more data augmentation 
during training (larger random data shifts - 64 vs. 40 and {\it mixup}).

Second, it can be seen that CNN has virtually identical
performance as PBN-DA-HMM (29 versus 29.3 errors).  
This is remarkable, first because PBN-DA-HMM used less data augmentation than CNN,
but more significantly, because PBN-DA-HMM is a generative
conditional  distributions classifier (See Section \ref{topsec}).
This is no doubt a result of the incorporation of both generative and discriminative approaches
in a single network.

\subsection{Combined Results}
It is well-known that combining the output of several classifiers usually 
improves the classification performance, especially if the combined 
classifiers (a) have comparable performance, and (b) they are based
on different methods or views of the data. When looking at the individual
performances in Table \ref{tab1r}, it is clear that the stage is set for good
classifier combination.  

\begin{figure}[h!]
\begin{center}
	\includegraphics[width=3.4in, height=3.3in]{ 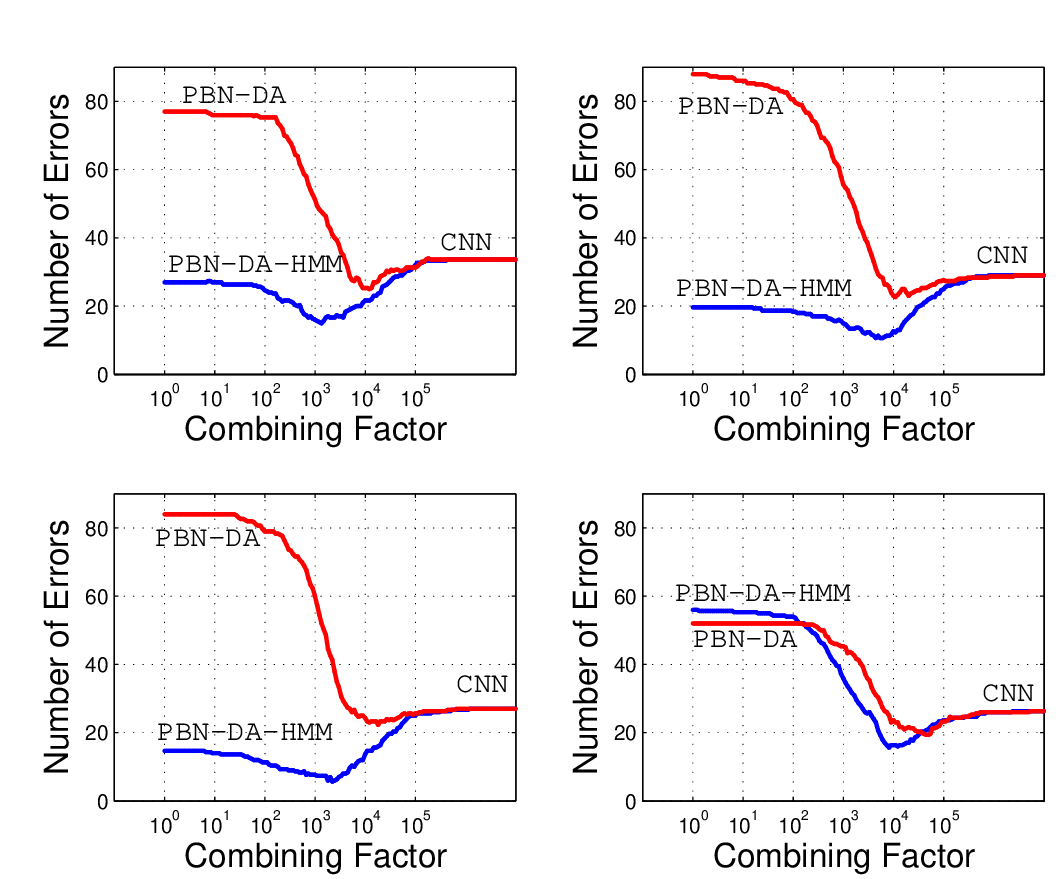}
	\caption{Mean number of classification errors out of 230 samples, for each of the four partitions when combining PBN-DA-HMM with CNN.}
	\label{figcomb4}
\end{center}
\end{figure}

Classifier combination (ensembling) results are shown as a function of the linear combination factor
for CNN with PBN-DA (in red) and CNN with PBN-DA-HMM (in blue) in Figure \ref{figcomb4}, and are
summarized in Table \ref{tab1r}.  Some interesting things to note are that (a) PBN-DA-HMM
works much better than PBN-DA, and combines also much better with CNN, (b) 
PBN-DA-HMM performs better than CNN in three of the four folds, 
and (c), that despite significantly worse performance
in the last fold, the combination still works exceptionally well.
This is an indication of the independence of the information being combined.
It can also be seen in the figure that the linear combination factor 
at which the minimum number of errors is achieved is about the same for
all four folds.  

\begin{figure}[h!]
\begin{center}
	\includegraphics[width=2.0in, height=2.0in]{ 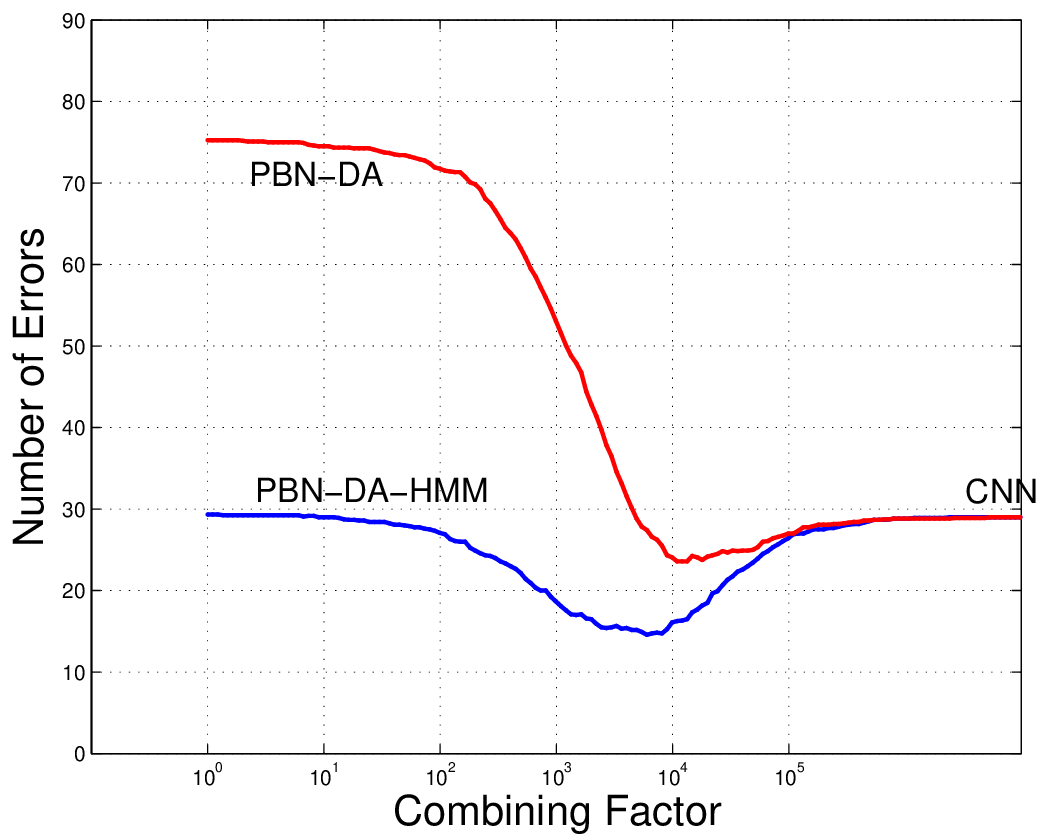}
	\caption{Mean number of classification errors out of 230 samples,
       averaged over the four folds when combining PBN-DA-HMM with CNN.}
	\label{figcomb}
\end{center}
\end{figure}

In Figure \ref{figcomb}, the average of the four folds is seen,
showing a factor of two reduction in classification errors at the best point.
In Table \ref{tab1r}, we summarize
the classifier performance obtained by linear combination of 
\ac{cnn} with PBN-DA and with PBN-DA-HMM, which is 
obtained at the optimal combining factor.  This factor, about 6000, 
is held constant across the data folds\footnote{The factor 6000 seems high, 
but this does not indicate anything about the relative importance of the two
statistics. It is only due to the wide difference in the scale of the
output statistics - log likelihood for PBN and cross-entropy
classifier cost for CNN.}.


\subsection{Reproducibility}
Because we used a non-standard subset of ESC50 data, and non-standard data partitions,
we make the feature data available, as well as software and instructions
to reproduce the results in this paper as \cite{PBNTk}.

\section{Experiment 2: Acoustic Trends Blue Fin Data Set}

\subsection{Data Description}
\begin{figure}[h!]
  \begin{center}
    \includegraphics[width=1.1in]{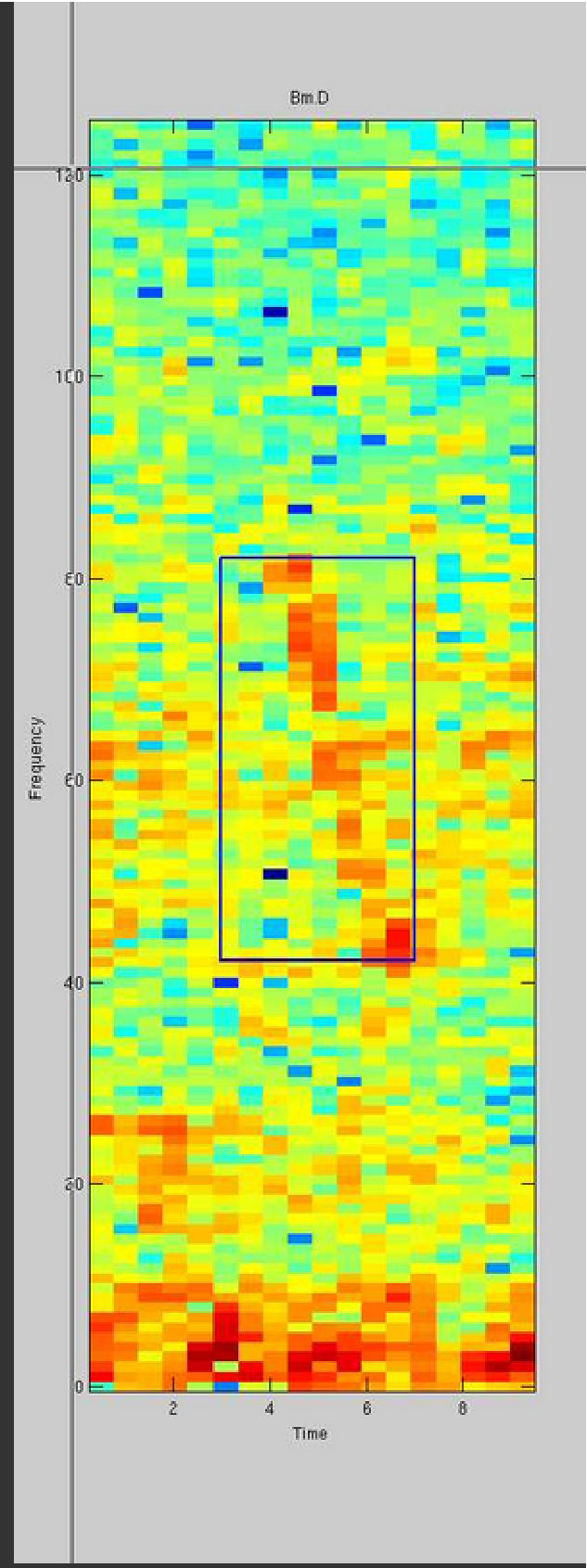}
    \includegraphics[width=1.1in]{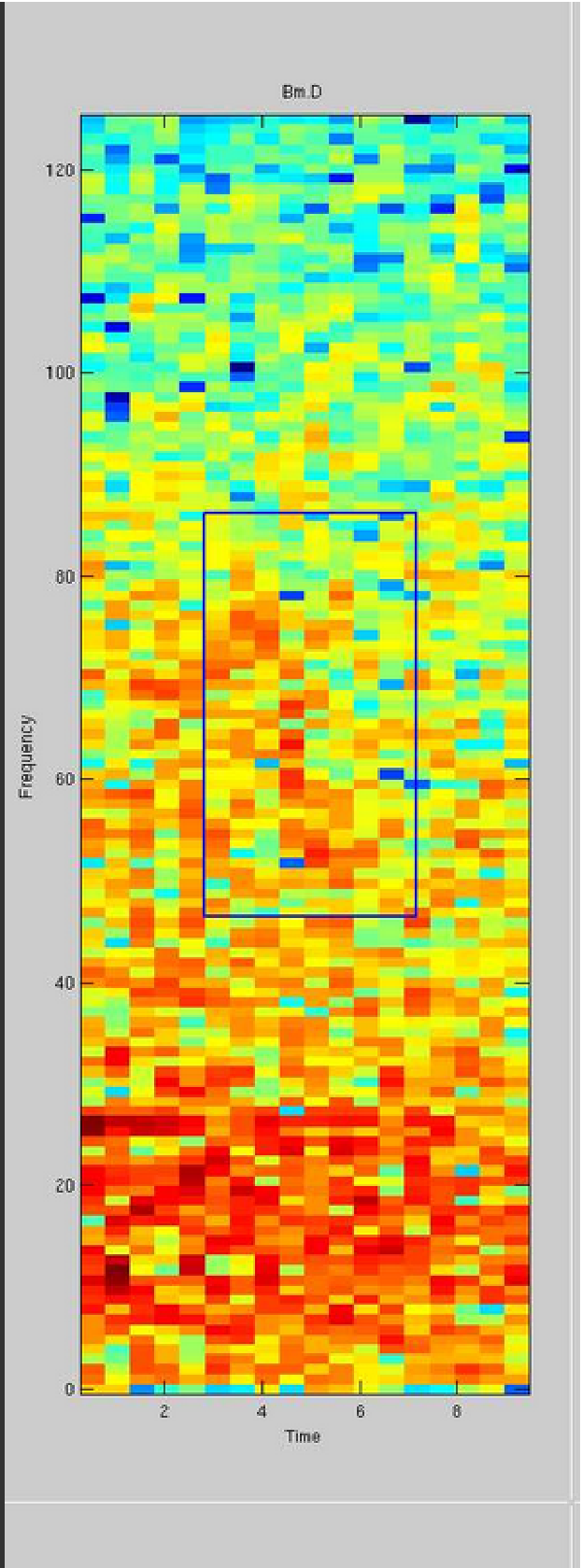}
    \includegraphics[width=1.1in]{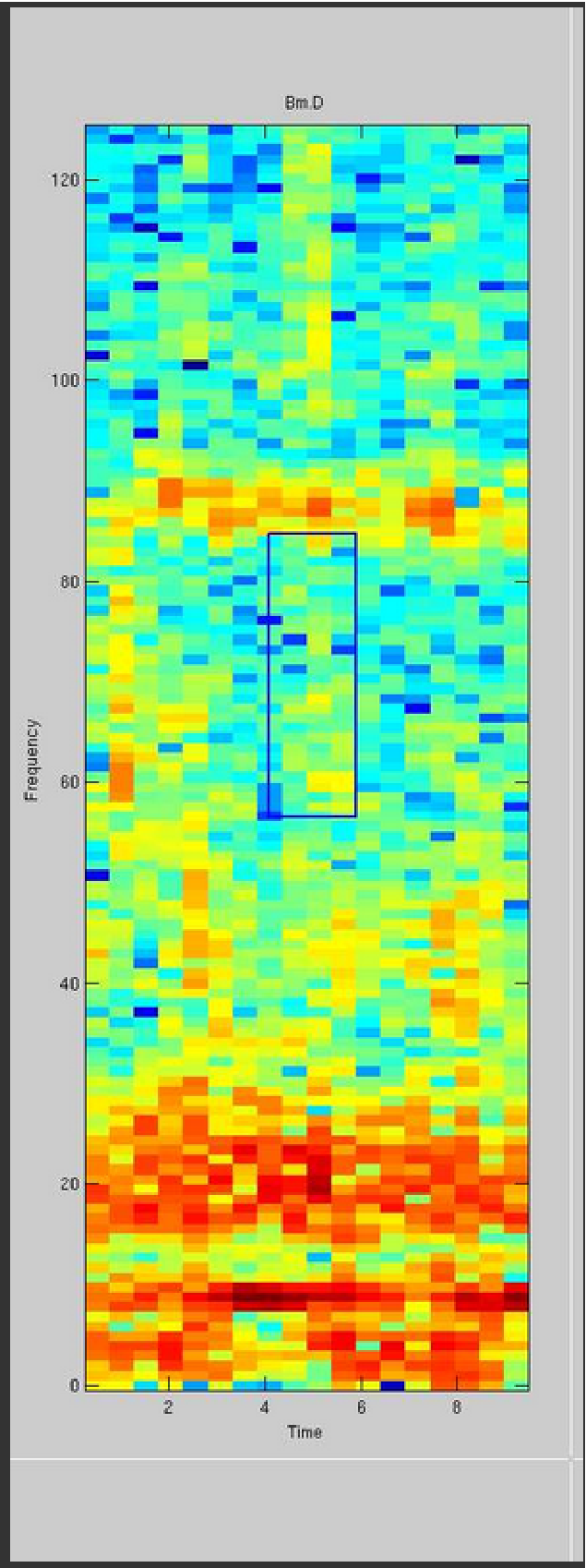}
  \caption{Examples of annotations from Acoustic Trends Blue Fin Data Set
of type ``Bp-Downsweep".}
  \label{aus6ex}
  \end{center}
\end{figure}

The Australian Acoustic Trends Blue Fin data set \cite{AusBluefin}
consists of acoustic recordings from various hydrophones, along with
a set of annotations of marine mammal vocalizations that describe the bounding boxes of each vocalization (start and end time, as well as start and end frequency).
We collected 200 examples of from each of six call types, denoted by
``BM-Ant-A ", ``BM.Ant-B ", ``BM.Ant-Z ", ``Bm.D ", ``Bp-20Hz", ``Bp-Downsweep".
Example annotations are shown in Figure \ref{aus6ex}
from class ``Bp-Downsweep".
An attempt was made to select only those annotations
that were recognizeable. The first of the three
samples in the figure was deemed recognizable, and the last two 
were not and were discarded.
In order to achieve 200 samples per event, it was necessary to
relax the criteria somewhat.  The data set is ``dirty" in the sense that 
many examples are weak and occur simultaneously with 
interfering noise and interfering calls.
A uniformly-sized time-window of 3072 samples at 250 Hz sample rate
(12 seconds) was extracted for each selected annotation.
For the current experiments, the time-series were converted to log-band energy features with the following parameters: FFT size 384 with 128-sample shift (2/3 overlap), 40 linear-spaced hanning-weighted frequency bands, resulting in a 
$40\times 24$ feature map per event.
Extracted .wav files and feature values
for the selected annotations have been made available
online \cite{CSFaus6}.

\subsection{Networks}
For the PBNs, a 5-layer network was used, consisting
of two convolutional layers, followed by three dense layers of
128, 32, and 6 nodes.
The first convolutional layer had 8 $7\times 13$ kernels
with $2\times 4$ downsampling, resulting in 8 output maps of
$12\times 10$ (dimensions always given in time$\times$freq).
The second convolutional layer had 48 $4\times 10$ kernels
with $2\times 1$ downsampling, resulting in 48 output maps of
$5\times 1$ (dimensions always given in time$\times$freq).
Linear activation was used at the output of the convolutional
layers, and truncated Gaussian (TG) activation function
was used at the output of all dense layers (see Table \ref{tab1a}). 
One of the primary reasons for the high computational
load of PBN is the need to invert matrices
of dimenson $M\times M$, where $M$ is the total output
dimension of a network layer.
One of the means of mitigating this 
is to use linear activation functions in a few of the first
network layers. These layers can then be grouped together
in what is called a {\it Gaussian group} 
with a low output dimension \cite{BagPBNHidim}.
We formed a Gaussian group from the first three layers.

For the benchmark CNN, the same network structure was used, but
max-pooling was used instead of downsampling, and
TG activation functions were used at the outputs of all layers.

\subsection{Training}
Six separate PBNs were trained (one on each data class)
using discriminative alignment and data augmentation of $+/- 3$ samples random 
circular time shifts and  $+/- 1$ sample random
frequency shift. Vernier shifts (i.e. not
quantized to integer shifts) were accomplished by 
shifting the data in the frequency domain using 2D-FFT.
 The self-combination factor $C=3$
was used during training (see Section \ref{selfcomb_sec}).
For testing and evaluation, the value $C=300$ was used.
The cross-entropy classifier cost (against ``all other classes")
was multiplied by 2000 before being subtracted from the
PBN log-likelihood value.
A learning rate of $2e^{-5}$ was used for 1500 epochs
or until there were no more training errors
(in the target class vs ``all other classes" FFNN experiment).

For the benchmark CNN, we used drop-out regularization as well.

\subsection{PBN-DA-HMM}
To implement PBN-DA-HMM, we tapped the output of the
second convolutional layer, making a $5\times 48$ feature map, 
where the first dimension was time.
The PDF of this feature was estimated using a 
4-state HMM, where each state was represented by a Gaussian
mixture (GMM) of 3 components. To prevent
covariance collapse, a value of 0.12
was added to the diagonal elements of the GMM covariance matrices.

\subsection{Results}
\begin{figure}[h!]
  \begin{center}
    \includegraphics[width=3.5in]{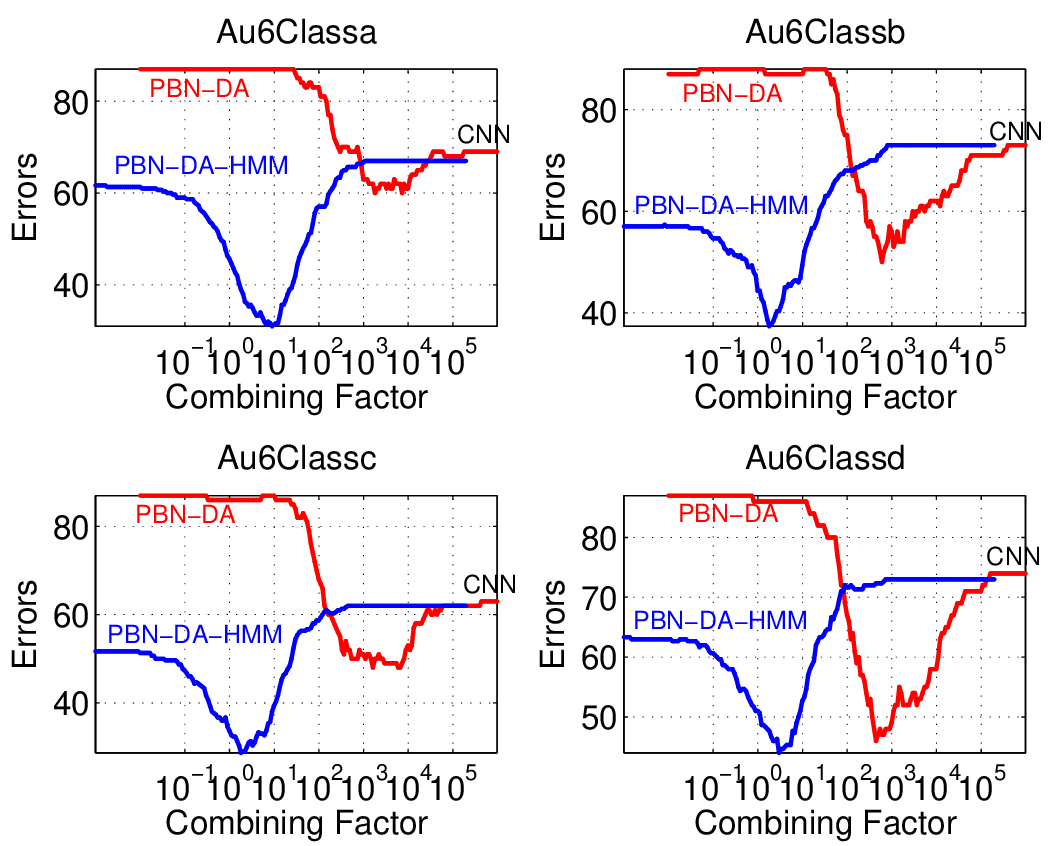}
  \caption{Results of
PBN-DA vs. PBN-DA-HMM in combination with CNN 
for Australian Blue Fin data. Shown is number of
errors out of 300 for each of the four data folds.  }
  \label{comb_au6hmm_all}
  \end{center}
\end{figure}

\begin{figure}[h!]
  \begin{center}
    \includegraphics[width=2.0in]{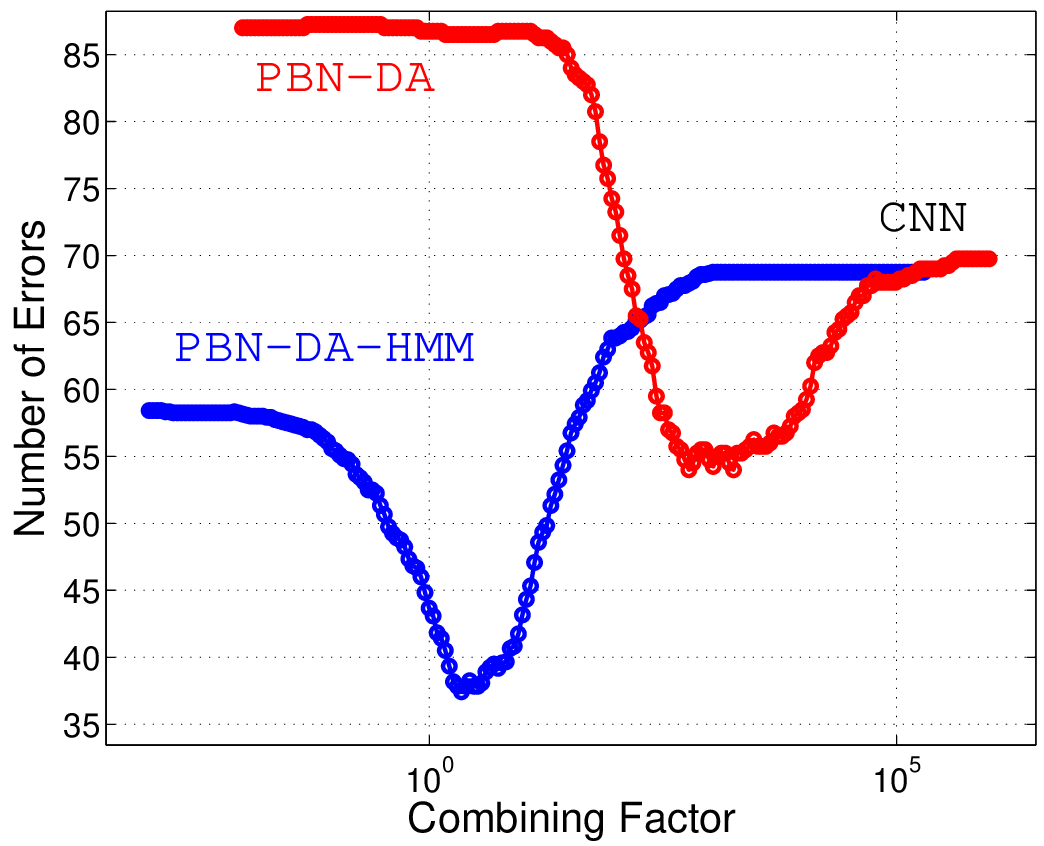}
  \caption{Results of PBN-DA vs. PBN-DA-HMM
in combination with CNN for Australian Blue Fin data.
Shown is number of errors out of 300 
averaged over the four data folds.}
  \label{comb_au6hmm_mean}
  \end{center}
\end{figure}

Results are shown for PBN-DA (in red) and PBN-DA-HMM (in blue)
in combination with the benchmark CNN for all four data
folds in Figure \ref{comb_au6hmm_all}, and averaged over the folds
in Figure \ref{comb_au6hmm_mean}.
Shown is total errors out of a total of 300 for each data fold.
These results are strikingly similar to 
the first experiment, using  a completely different data set.
As in the first experiment, PBN-DA performed well, but not as well
as the benchmark CNN, and resulted in significant error reduction when combined
with the CNN. Also, as in the first experiment, PBN-DA-HMM
 performed exceedingly well, better than 
the benchmark CNN, and resulted in a very significant 
2:1 error reduction.

\section{Conclusion and Future Work}
In this paper, generative classifiers called PBN-DA and PBN-DA-HMM
have been described that combine
both generative and discriminative methods in a single network. 
Generative PBN networks are trained 
separately on each class using discriminative alignment, ensuring that the 
generative models are selective against the other data classes. 
Optionally, when (a) the input data has a time dimension 
where convolutional layers can be applied, and (b) 
a suitable reduced-dimension latent variable is present
at the output of an intermediate network layer, then 
the PBN-DA-HMM method can be applied.  To do this, the PBNs 
are shortened, tapping off at the suitable intermediate layer.
The probability density of these latent variables is estimated using HMMs,
which leverage the Markov assumption to create good probability density density estimates.  When a time-dimension is not avaibale, it should
be possible to apply Gaussian mixture model effectively.

Two separate experiments using vastly different acoustic data
sets, confirm that the approach does indeed attain 
some the best qualities of both discriminative and
generative approaches.  In fact, it is seen that
the performance of PBN-DA-HMM is virtually identical
to or better than a state of the art CNN, and by linear combining with CNN
attains a factor of two reduction in error rate.  The relatively high computational burden of PBN
means that the approach is suited to problems with high
cost of errors, such as in military and human safety applications.

\par
Software implementations of PBN and related programs \cite{CSFTk}, \cite{PBNTk} and data sets for the current problems are made available online \cite{CSFaus6}.

\par
For future work, it is planned to extend the results to the full ESC50 data set.

\bibliographystyle{IEEEtran}
\bibliography{ppt}
\end{document}

\appendix[ Appendix \ref{app1}: Segmentation of Time-Series]
%\section{Segmentation of Time-Series}
\label{app1}
Classification of acoustic events is an important task in machine learning.
A very important decision in the pre-processing of acoustic
events analysis is the selection of segmentation window size,
weighting function, and overlap.
A compromise must be struck between the desire to capture short-duration
character versus log-duration narrow-band signatures.
However, when applying PDF projection in the manner explained above, to
allow the integration of multiple choices of segmentation
size into a ``mixture of models", the definition of
``input data" becomes unclear. 
When using overlapped window functions with weighting
such as Hanning (raised cosine) weighting, two adjacent data windows have a significant
overlap. Therefore, a single time-series sample appears more than
once in the set of windows.  Let
${\bf x} = \left\{ {\bf x}_1, {\bf x}_2, \ldots {\bf x}_K\right\}$ be the
collection of $K$ time windows. If we apply (\ref{ppt0}), then
due to the mutual dependence of the segments, $p_{x,0}({\bf x})$ 
is not equal to $\prod_{i=1}^K \; p_{x,0}({\bf x}_i).$
The statistical dependence may be complicated
and differs across the various segment sizes.
Therefore, comparing projected PDFs
$G_1(\bfx)$ and $G_2(\bfx)$ has no theoretical basis if
$G_1(\bfx)$ and $G_2(\bfx)$ are derived from different segmentation sizes.
Fortunately, when using Hanning weighting with  $2/3$ overlap,
called {\it hanning-3} segmentation,
a solution to the problem can be found \cite{BagHanning}.  
Let the time-series of
length $N$ be segmented using segment size $K_l$ and window time shift $S_l$,
having $O_l = K_l-S_l$ time samples of overlap. 
If we circularly-index the data
such that $x_{N+i}=x_i$, we will obtain exactly $T_l = N/S_l$ segments.

Note that we need to insure that $K_l$ is divisible by
3 and $N$ is divisible by $S_l$, for all $l$.
This is not a severe restriction.  For example, let $K_l$ be selected 
from the set  $[72, \; 96, \; 144, \; 192, \; 288, \; 384, \; 576, \; 768].$
If we truncate the time-series to a multiple of $N=768$ time samples, and with
$S_l=K_l/3$, we could always achieve these requirements.

Let the {\it hanning-3} segmentation $l$ be given by 
\beq
\begin{array}{rcl}
\bfx^{l} &= &
 \left\{ [x_1 w_1, x_2 w_2 \ldots x_{K_l} w_{K_l}], \;  \right. \\ \\
  &&  [x_{(S_l+1)} w_1, \ldots x_{(S_l+K_l)} w_{K_l}],   \\ \\
  && \left. [x_{(2S_l+1)} w_1, \ldots x_{(2S_l+K_l)} w_{K_l}], \ldots \right\}
\end{array}
\label{segm3}
\eeq
where $w_i, \; 1\leq i \leq K,$ are the Hanning-3 weights 
\beq
  w_i = \frac{\sqrt{2}}{3} \left[ 1+\cos\left(2\pi(i-1)\over K\right)\right].
 \label{wh3}
\eeq
Note that $\bfx^{l}$ has 
exactly $3N$ time samples, as opposed to the original data $\bfx$ that has $N$
(each time sample in $\bfx$ appears three times in $\bfx^{l}$,
with different weights).

To use various {\it hanning-3} segmentations together
in a class-specific classifier, we apply the concept
of {\it virtual input data.}
In Figure \ref{hann3}, we illustrate the Hanning-3
window functions with 2/3 overlap (right) and compare with
50\% overlap (left). Note that for Hanning-3,
both the sum of the window functions
(center graph) and the sum of the squares of the window functions
(bottom graph) are constant.
  \begin{figure}[h]
  \begin{center}
    \includegraphics[height=1.4in,width=1.7in]{hann3-b.eps}
    \includegraphics[height=1.4in,width=1.7in]{hann3-a.eps}
  \caption{ Illustration of the properties of 50\% overlapping
(left) and Hanning-3 window functions (right).}
  \label{hann3}
  \end{center}
  \end{figure}
This property does not hold for 50\% overlap,
but only for $2/3$, $3/4$, and higher overlap.
The property leads to the observation
that %In a previous publication \cite{BagHanning}, we showed that
two different hanning-3 segmentations 
$\bfx^{l}$ and $\bfx^{m}$ where $l\neq m$, are related
by an orthogonal linear transformation.
Specifically, it is shown \cite{BagHanning} that for any $l\neq m$,
there exists an ortho-normal matrix ${\bf U}_{l,m}$
such that $$\bfx^{l} = {\bf U}_{l,m} \; \bfx^{m},$$ 
where ${\bf U}_{l,m}$ has a determinant of 1.
We call $\bfx^{l}$  {\it virtual} input data because given any PDF defined for one
segmentation, we can use the change of variables theorem to find the PDF 
for another segmentation, and it is identical.
Therefore, any or all of the hanning-3 segmentations
can serve as the ``input data".
%This is illustrated in Figure \ref{hann23} in which
%the output of each segmentation operation is considered
%as the ``virtual input data" of each branch.  
%Each branch has a different virtual input data, but they are 
%considered ``equivalent".  The projected
%likelihood function for $\bfx^{l}$ may be compared
%to the projected the likelihood function for $\bfx^{m}$.
%The projected likelihood function for the
%virtual input data of each branch is formed
%%by adding the feature log-likelihood to the log J-function.  
%At the output, the branch log-likelihoods are
%compared or combined.
%  \begin{figure*}[h]
%  \begin{center}
%    \includegraphics[height=2.8in,width=6.0in]{hann23b.eps}
%  \caption{ Illustration of the means of calculating
%the log-likelihood for a given class hypothesis $k$
%from input data $\bfx$ using three processing branches,
%each using a different segmentation size.  The output of each 
%hanning-3 segmentation operation is the ``virtual input data" 
%of the branch.}
%  \label{hann23}
%  \end{center}
%  \end{figure*}

\bibliographystyle{/home/paul.baggenstoss/tex/IEEE/Bibtex/IEEEtran}
\bibliography{ppt}
\end{document}

%% file: macros.tex
\newcommand{\defined}{\stackrel{\mbox{\tiny$\Delta$}}{=}}
\newtheorem{example}{Example}
\newtheorem{conclusion}{Conclusion}
\newtheorem{assumption}{Assumption}
\newtheorem{definition}{Definition}
\newtheorem{problem}{Problem}
\newcommand{\erf}{{\rm erf}}

\newcommand{\sst}{\scriptstyle }
\newcommand{\xparen}{\mbox{\small$(\bfx)$}}
\newcommand{\hojz}{H_{0j}\mbox{\small$(\bfz)$}}
\newcommand{\Hozj}{H_{0,j}\mbox{\small$(\bfz_j)$}}
\newcommand{\smallmath}[1]{{\scriptstyle #1}}
\newcommand{\Hoz}[1]{H_0\mbox{\small$(#1)$}}
\newcommand{\Hozp}[1]{H_0^\prime\mbox{\small$(#1)$}}
\newcommand{\Hozpp}[1]{H_0^{\prime\prime}\mbox{\small$(#1)$}}
\newcommand{\hoz}{\Hoz{\bfz}}
\newcommand{\hooz}{\Hozp{\bfz}}
\newcommand{\hoooz}{\Hozpp{\bfz}}
\newcommand{\smJ}{{\scriptscriptstyle \! J}}
\newcommand{\smK}{{\scriptscriptstyle \! K}}

\newcommand{\erfc}{{\rm erfc}}
\newcommand{\bitem}{\begin{itemize}}
\newcommand{\dsum}{{ \displaystyle \sum}}
\newcommand{\eitem}{\end{itemize}}
\newcommand{\benum}{\begin{enumerate}}
\newcommand{\eenum}{\end{enumerate}}
\newcommand{\bdm}{\begin{displaymath}}
\newcommand{\bfzro}{{\underline{\bf 0}}}
\newcommand{\bfone}{{\underline{\bf 1}}}
\newcommand{\edm}{\end{displaymath}}
\newcommand{\beq}{\begin{equation}}
\newcommand{\bea}{\begin{eqnarray}}
\newcommand{\eea}{\end{eqnarray}}
\newcommand{\cali}{ {\cal \bf I}}
\newcommand{\caln}{ {\cal \bf N}}
\newcommand{\barray}{\begin{displaymath} \begin{array}{rcl}}
\newcommand{\earray}{\end{array}\end{displaymath}}
\newcommand{\eeq}{\end{equation}}
\newcommand{\qed}{\framebox{$\;$}}
\newcommand{\btheta}{\mbox{\boldmath $\theta$}}
\newcommand{\bTheta}{\mbox{\boldmath $\Theta$}}
\newcommand{\blam}{\mbox{\boldmath $\Lambda$}}
\newcommand{\bdelta}{\mbox{\boldmath $\delta$}}
\newcommand{\bgamma}{\mbox{\boldmath $\gamma$}}
\newcommand{\balpha}{\mbox{\boldmath $\alpha$}}
\newcommand{\bbeta}{\mbox{\boldmath $\beta$}}
\newcommand{\balphascript}{\mbox{\boldmath ${\scriptstyle \alpha}$}}
\newcommand{\bbetascript}{\mbox{\boldmath ${\scriptstyle \beta}$}}
\newcommand{\bLambda}{\mbox{\boldmath $\Lambda$}}
\newcommand{\bDelta}{\mbox{\boldmath $\Delta$}}
\newcommand{\bomega}{\mbox{\boldmath $\omega$}}
\newcommand{\bOmega}{\mbox{\boldmath $\Omega$}}
\newcommand{\blambda}{\mbox{\boldmath $\lambda$}}
\newcommand{\bphi}{\mbox{\boldmath $\phi$}}
\newcommand{\bpi}{\mbox{\boldmath $\pi$}}
\newcommand{\bnu}{\mbox{\boldmath $\nu$}}
\newcommand{\brho}{\mbox{\boldmath $\rho$}}
\newcommand{\bmu}{\mbox{\boldmath $\mu$}}
\newcommand{\sigi}{\mbox{\boldmath $\Sigma$}_i}
\newcommand{\bfu}{{\bf u}}
\newcommand{\bfx}{{\bf x}}
\newcommand{\bfb}{{\bf b}}
\newcommand{\bfk}{{\bf k}}
\newcommand{\bfc}{{\bf c}}
\newcommand{\bfv}{{\bf v}}
\newcommand{\bfn}{{\bf n}}
\newcommand{\bfK}{{\bf K}}
\newcommand{\bfh}{{\bf h}}
\newcommand{\bff}{{\bf f}}
\newcommand{\bfg}{{\bf g}}
\newcommand{\bfe}{{\bf e}}
\newcommand{\bfr}{{\bf r}}
\newcommand{\bfw}{{\bf w}}
\newcommand{\calX}{{\cal X}}
\newcommand{\calZ}{{\cal Z}}
\newcommand{\bx}{{\bf x}}
\newcommand{\bb}{{\bf b}}
\newcommand{\by}{{\bf y}}
\newcommand{\bfy}{{\bf y}}
\newcommand{\bfz}{{\bf z}}
\newcommand{\bfs}{{\bf s}}
\newcommand{\bfa}{{\bf a}}
\newcommand{\bfA}{{\bf A}}
\newcommand{\bfB}{{\bf B}}
\newcommand{\bfV}{{\bf V}}
\newcommand{\bfZ}{{\bf Z}}
\newcommand{\bfH}{{\bf H}}
\newcommand{\bfX}{{\bf X}}
\newcommand{\bfR}{{\bf R}}
\newcommand{\bfF}{{\bf F}}
\newcommand{\bfS}{{\bf S}}
\newcommand{\bfC}{{\bf C}}
\newcommand{\bfI}{{\bf I}}
\newcommand{\bfO}{{\bf O}}
\newcommand{\bfU}{{\bf U}}
\newcommand{\bfD}{{\bf D}}
\newcommand{\bfY}{{\bf Y}}
\newcommand{\bSig}{{\bf \Sigma}}
\newcommand{\test}{\stackrel{<}{>}}
\newcommand{\zmk}{{\bf Z}_{m,k}}
\newcommand{\zlk}{{\bf Z}_{l,k}}
\newcommand{\zm}{{\bf Z}_{m}}
\newcommand{\ssq}{\sigma^{2}}
\newcommand{\dint}{{\displaystyle \int}}
\newcommand{\ds}{\displaystyle }
\newtheorem{theorem}{Theorem}
\newcommand{\postscript}[2]{ \begin{center}
    \includegraphics*[width=3.5in,height=#1]{#2.eps}
    \end{center} }